\documentclass[twoside]{article}

\usepackage[accepted]{aistats2025}

\usepackage[round]{natbib}

\usepackage[utf8]{inputenc} 
\usepackage[T1]{fontenc}    
\usepackage{hyperref}       
\usepackage{url}            
\usepackage{booktabs}       
\usepackage{amsfonts}       
\usepackage{nicefrac}       
\usepackage{microtype}      
\usepackage{xcolor}         
\usepackage{amsmath,amssymb,amsfonts}
\usepackage{mathtools}
\usepackage{dsfont}
\usepackage{wrapfig}
\usepackage{xr}
\usepackage{subcaption}
\usepackage{pifont}
\usepackage{algorithm}
\usepackage{algpseudocode}
\newtheorem{lemma}{Lemma}
\newtheorem{Proposition}{Proposition}
\newtheorem{problem}{Problem}
\input{mysymbol.sty}

\newcommand{\cmark}{\ding{51}}%
\newcommand{\xmark}{\ding{55}}%

\begin{document}
%

%

\twocolumn[

\aistatstitle{Scalable Implicit Graphon Learning}

\aistatsauthor{ Ali Azizpour \And Nicolas Zilberstein \And Santiago Segarra }

\aistatsaddress{ Rice University, USA \And  Rice University, USA \And Rice University, USA } ]

\begin{abstract}
Graphons are continuous models that represent the structure of graphs and allow the generation of graphs of varying sizes. 
We propose Scalable Implicit Graphon Learning (SIGL), a scalable method that combines implicit neural representations (INRs) and graph neural networks (GNNs) to estimate a graphon from observed graphs. 
Unlike existing methods, which face important limitations like fixed resolution and scalability issues, SIGL learns a continuous graphon at arbitrary resolutions. 
GNNs are used to determine the correct node ordering, improving graph alignment.
Furthermore, we characterize the asymptotic consistency of our estimator, showing that more expressive INRs and GNNs lead to consistent estimators. 
We evaluate SIGL in synthetic and real-world graphs, showing that it outperforms existing methods and scales effectively to larger graphs, making it ideal for tasks like graph data augmentation.
An implementation of SIGL, along with the code to reproduce our results, can be found at our {GitHub repository} (\url{https://github.com/aliaaz99/SIGL}).
\end{abstract}

\section{INTRODUCTION}
\label{sec:intro}

Graphs are powerful tools for visualizing complex relationships between various objects.
For instance, we can examine the relationships between users in a social~\citep{socialnetwork} or wireless network~\citep{wireless}, or represent interactions within brain recordings~\citep{braingraph} or various genomic sequences~\citep{komb, grassrep}.
In many cases, we have access to a collection of graphs that share \emph{structural properties}. 
Instead of analyzing each graph individually, it is beneficial to identify a common underlying model from which these graphs originate. 
We can use this model to devise solutions that can be then applied to any instance of a graph drawn from this common model.

Graphons~\citep{lovasz2012large, diaconis2007graph} are well-suited for representing a common underlying graph model.
A graphon can be interpreted either as the limit object of a convergent sequence of graphs or as a non-parametric model for generating graphs.
Despite their success in a plethora of applications~\citep{parise2023graphon, navarro2022joint,gao2019graphon, roddenberry2021network}, graphons are typically unknown in real-world applications and need to be estimated~\citep{ruiz2021graphon}.
This demands \emph{scalable estimation techniques} that can deal with graphs of varying sizes.

In this work, we address the problem of estimating the graphon underlying multiple (potentially large-scale) networks sampled from the same (unknown) random graph model.
Graphon estimation from multiple observed graphs is a longstanding problem~\citep{sas, ignr, sassbm, keshavan2010matrix, gwb, usvt}.
Still, existing methods have drawbacks such as fixed resolution and unknown sorting of graphs with respect to the graphon. 
These methods often treat the graphon as a fixed-resolution matrix, limiting the range of graphon functions they can capture. 
Additionally, relying on predefined sorting metrics like node degree or k-core can fail to differentiate nodes with similar characteristics. 
To address these challenges, IGNR~\citep{ignr} uses implicit neural representation (INR)~\citep{siren} to obtain a continuous representation of the graphon and uses the Gromov-Wasserstein (GW) distance as a loss function. 
Although the GW distance has been applied in previous works~\citep{xu2019scalable, gwb}, the combinatorial nature of GW yields an optimization problem that does not \emph{scale} when learning from large graphs.
Moreover, current approaches typically focus on learning a single graphon, despite the need to model varying graphons over latent parameters like time.

\begin{figure*}[!th]
    \centering
    \includegraphics[width=0.9\textwidth]{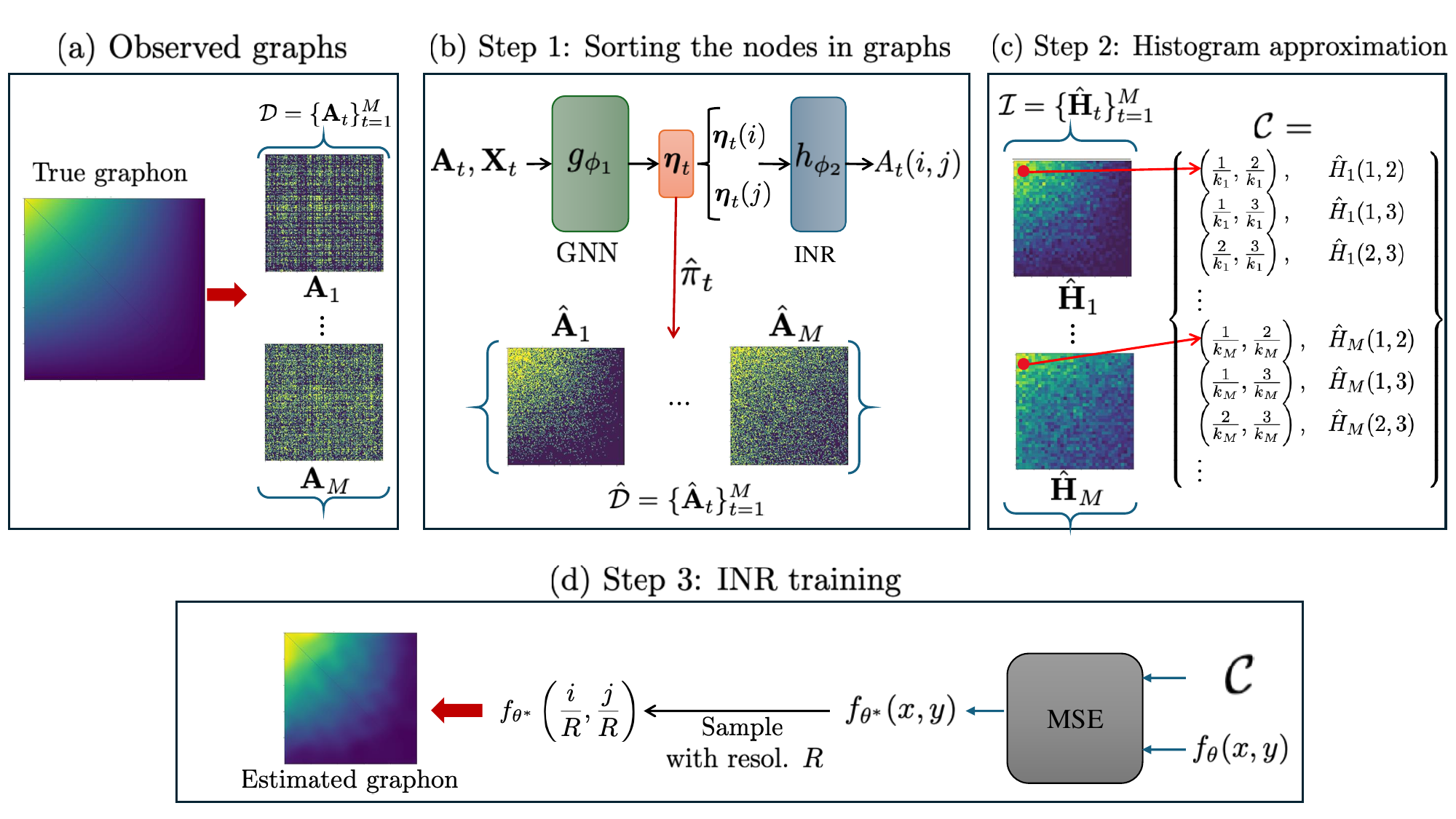}
    \caption{Overview of SIGL.
    (a)~A set of graphs is observed from the true (unknown) graphon.
    (b)~Step 1: A GNN-INR module learns the latent variables of the nodes in the observed graphs, enabling alignment of the graphs.
    (c)~Step 2: We build the dataset $\ccalC$ containing the upper triangular part the histogram approximations in the dataset $\ccalI$. 
    (d)~Step 3: An INR is trained using the coordinates and histogram values to estimate the graphon. The learned INR can be sampled at any arbitrary resolution $R$ to generate the final estimated graphon.
    }
    \label{fig:overview}
\end{figure*}

To address these challenges, we propose Scalable Implicit Graphon Learning (SIGL), which combines implicit neural representations (INRs)\citep{siren} to parametrize the graphon, and graph neural networks (GNNs)\citep{gnnsurvey} to learn a node ordering. 
Like IGNR~\citep{ignr}, we use an INR to represent the graphon.
However, a key difference is that we optimize the MSE instead of the GW distance.
This requires aligning the observed graphs with the learnable graphon based on an unknown node ordering. 
To handle this, we leverage GNNs to learn distinctive node features and estimate the true ordering, resulting in a scalable and computationally efficient approach for large, real-world graphs.
An overview of our method is illustrated in Fig.~\ref{fig:overview}.

We validate the advantages of SIGL through extensive experiments on both synthetic and real-world graphs. 
Our synthetic experiments show that SIGL achieves significant scalability compared to methods that rely on the GW distance, which struggles with larger graphs. 
Additionally, we apply SIGL in the context of graphon mixup~\citep{gmixup, navarro2023graphmad} for graph data augmentation, where our method enhances the performance of downstream tasks such as graph classification. 
In comparison to existing methods, SIGL consistently improves classification accuracy, illustrating its capacity to effectively capture the structural properties of graphs and boost task performance in real-world applications.

In summary, our main contributions are as follows:

\noindent 
1) We propose \textbf{Scalable Implicit Graphon Learning} (SIGL), a \emph{scalable training framework} for learning resolution-free parameterizations (INRs) of graphons.

\vspace{0.5mm}
\noindent 
2) We introduce a method to align observed graphs by estimating the true node ordering using a GNN, instead of relying on predefined metrics.

\vspace{0.5mm}
\noindent 
3) We formally characterize the \emph{asymptotic consistency} of the
estimator obtained with SIGL as a function of the representation capabilities of the used INR. 

\vspace{0.5mm}
\noindent 
4) Through extensive numerical experiments, we demonstrate SIGL's scalability and superior performance compared to the state of the art.
Additionally, we apply our method to graph mixup, highlighting its effectiveness in solving real-world problems.

\noindent\textbf{Related works.}
Classical graphon estimation methods such as stochastic block approximation ({SBA})~\citep{sba}, smoothing-and-sorting ({SAS})~\citep{sas}, largest gap ({LG})~\cite{LG}, and universal singular value thresholding ({USVT})~\citep{usvt} typically require aligning the observed graphs based on a predefined metric before estimating a fixed-resolution matrix as the underlying graphon.
For instance, the aforementioned methods sort the observed graphs by empirical degree; this sorting is motivated by the fact that the degree function of a graphon (as well as other centrality measures) arises as the limit of its counterpart measure in sequences of increasingly large graphs~\citep{avella2018centrality}.
SAS (which is an improved variant of SBA) then computes histogram approximations of the sorted adjacency graphs and smooths the approximated histogram using total variation minimization.
On the other hand, USVT~\citep{usvt} is a matrix completion method that exploits the low-rank structure of many networks by thresholding the singular values to retain those above a certain threshold.
Both approaches produce graphons with resolutions limited by the sizes of the observed graphs. 
Also, when the nodes of the graphs have similar characteristics with respect to the predefined metric (such as similar degree), these methods fail to align the graphs accurately.

To address the alignment problem, {GWB}~\citep{gwb} and implicit graphon neural representation ({IGNR})~\citep{ignr} use the GW distance to align the graphs. 
GWB estimates the graphon by computing the GW barycenter of the observed graphs, though it still produces a fixed-resolution estimate and then upsamples it to the desired resolution.
IGNR uses INRs to estimate a resolution-free representation of the true graphon from a set of observed graphs. 
It directly feeds the observed graphs into its model and uses the GW distance between the reconstructed graphs from the estimated graphon and the observed graphs as its loss function.
However, the reliance on the GW distance in both methods makes them impractical for larger graphs.


\vspace{-0.1in}
\section{BACKGROUND}
\label{sec:background}
\vspace{-0.1in}

\paragraph{Graphon.}
A graphon is a bounded symmetric measurable function $\omega : [0,1]^2 \rightarrow [0,1]$. 
Its domain can be interpreted as the edges in an infinitely large adjacency matrix, while the range of $\omega$ represents edge probabilities. 
By this definition, a graphon serves as a \emph{random graph generative model from which graphs with similar structural characteristics can be sampled.}
Generating an undirected graph $G = (\ccalV, \ccalE)$ of $n$ nodes from a graphon $\omega$ involves two steps: (1) selecting a random value between 0 and 1 for each node as its \emph{latent variable}, and (2) assigning an edge between nodes with a probability equal to the value of the graphon at their randomly sampled latent variables. 
Formally, the steps are as follows:
\begin{align}
\label{eq:stochastic_sampling}
    \eta _i &\sim \operatorname{Uniform} ([0,1]),\quad \forall \; i=1,\cdots,n, \\ \nonumber
    A(i,j) &\sim \text{Bernoulli}\left(\omega(\eta_i, \eta_j)\right),\quad \forall\; i,j=1,\cdots,n, 
\end{align}
where the latent variables $\eta_i \in [0,1]$ are independently drawn for each node $i$, and $\bbA \in \{0,1\}^{n\times n}$ is the adjacency matrix of $G$.

\paragraph{INR.}
In the INR setting, the observed data $\mathbf{s}_i \in \mathbb{R}^p$ are considered as discrete realizations $\mathbf{s}_i = f(\mathbf{x}_i)$ of some unknown signal $f: \mathcal{X} \to \mathcal{S}$ (with $\mathcal{X} \subseteq \mathbb{R}^d$ and $\mathcal{S} \subseteq \mathbb{R}^p$) sampled at coordinates $\mathbf{x}_i \in \mathcal{X}$ for $i=1, \ldots, N$. 
For example, in the context of an image, $\mathbf{x}_i = (x_i, y_i)$ represents a single pixel coordinate, and $\mathbf{s}_i$ is the 3-dimensional RGB value at that pixel.
So, the unknown function $f$ is implicitly defined through $\mathbf{x}_i$'s and $\mathbf{s}_i$'s. 
An INR is a neural architecture $f_\theta: \mathcal{X} \to \mathcal{S}$, typically a multilayer perceptron (MLP) parameterized by $\theta$.
This network is trained on the pairs $(\mathbf{x}_i, f(\mathbf{x}_i))$ to approximate $f$. 
Since $f_\theta$ is trained on the full continuous domain of $\mathcal{X}$, it is resolution-free, and at inference time, we can approximate the signal value at arbitrary points in $\mathcal{X}$ by evaluating $f_\theta$ at that specific point.

\section{SCALABLE LEARNING OF IMPLICIT GRAPHON NEURAL REPRESENTATIONS}\label{sec:Methods}

In Section~\ref{subsec:single_graphon}, we introduce SIGL, our proposed method for efficiently learning a graphon parametrized by an INR from a set of observed graphs.
We characterize the asymptotic consistency of the estimator obtained with SIGL in Section~\ref{subsec:theory}.

\subsection{Scalable Learning of a Graphon}
\label{subsec:single_graphon}
\vspace{-0.05in}

We consider an \emph{unknown} graphon $\omega$.
While we do not have access to $\omega$, we observe a set of graphs $\ccalD = \{\bbA_t\}_{t=1}^{M}$ generated from $\omega$.
In this setting, our problem is defined as follows:
\vspace{-0.05in}
\begin{problem}\label{prob1}
Given a set of graphs \( \ccalD = \{\bbA_t\}_{t=1}^{M} \) drawn from an unknown graphon \( \omega \), estimate \( \omega \).
\end{problem}
\vspace{-0.05in}
To solve Problem \ref{prob1}, we represent the graphon using an INR \( f_\theta: [0,1]^2 \to [0,1] \) parametrized by \( \theta \). 
As explained in Section~\ref{sec:intro}, learning a graphon INR from large-scale graphs is challenging: using the GW distance as a loss function leads to a learning process that does not \emph{scale}.
To circumvent this, SIGL learns the parameters \( \theta \) by following a three-step procedure.
First, we sort the observed graphs by estimating the latent variables of the nodes using a GNN structure along and an auxiliary graphon, which helps align the graphs. 
Next, we approximate the graphon by constructing a histogram from the sorted adjacency matrix. 
Finally, we learn a parametric representation of the graphon using an INR.
An overview of the steps involved in SIGL is illustrated in Fig.~\ref{fig:overview}.
A detailed algorithm is also included in Appendix~\ref{sec:appendix_str}.
We now expand on each step.

\noindent\textbf{Step 1: Sorting the nodes in the observed graphs.}
This step aims to replicate the graph generation process described in Section~\ref{sec:background}.
Intuitively, we \emph{rearrange the nodes of the observed graphs based on a common node ordering, ensuring they are well-aligned according to the same underlying graphon.}
This alignment is the essential first step for accurately estimating the shared graphon structure. 
However, \emph{the common node ordering of the observed graphs is unknown}, as they are labeled with arbitrary indexing --- there is no relation between a node's index and its underlying latent variable $\eta_i$ in~\eqref{eq:stochastic_sampling}. 
Previous approaches have relied on predefined metrics such as empirical node degree to sort (and re-index) the nodes. 
However, these metrics may fail to distinguish nodes when they have similar characteristics. 
For instance, in stochastic block model (SBM) graphs, nodes in different communities may share the same expected degree, leading to a uniform sorting that does not capture the necessary node distinctions.

To address this issue, we propose sorting the nodes by estimating their corresponding \emph{unknown} latent variables. 
Specifically, we learn the latent variables \(\hbeta_{t} = \{\hat{\eta}_i\}_{i=1}^{n_t}\) associated to each node $i$ of graph $G_t$ with $n_t$ nodes.
In addition, to ensure that all observed graphs follow the same sorting given by $\omega$, we learn an auxiliary graphon $h$.
In other words, given the adjacency matrix $\bbA_{t}$ of graph $G_t$, we have \(h(\hat{\eta}_{i,t},\hat{\eta}_{j, t})=A_{t}(i,j)\).

To learn the latent variables $\hbeta_t$ and the corresponding sorting, we propose a neural architecture consisting of two components associated with the $(i)$ latent variables and $(ii)$ the auxiliary graphon, respectively.
The first component is a GNN \( g_{\phi_1} \), whose outputs consist of the latent variables, i.e., \(\hbeta_t = g_{\phi_1}(\bbA_t, \bbX_t)\), where \(\bbX_t\) represents node features, here defined as \(\bbX_t \sim \mathcal{N}(0,1)\).
The second component is an INR, representing the auxiliary graphon $h_{\phi_2}$, and parametrized by $\phi_2$.
Combining both components yields a common underlying graphon, mapping the estimated latent variables to the edge between two nodes, which serves as a proxy for learning the underlying sorting.
We learn the parameters $\phi=\{\phi_1,\phi_2 \}$ by minimizing the loss function in~\eqref{eq:loss0}, which compares the predicted edges with the true edges between each pair of nodes $(i,j)$.
\begin{align}\label{eq:loss0}
    \ccalL(\phi) = \sum_{t=1}^{M} \frac{1}{n_t^2}\sum_{i,j=1}^{n_t} \left[ A_t(i,j) - h_{\phi_{2}}\left(\hat{\eta}_{t}(i),\hat{\eta}_{t}(j)\right)\right]^2.
\end{align}
Finally, we define a permutation $\hat{\pi}$ based on the estimated latent variables of the nodes such that $\hbeta_t(\hat{\pi}(1)) \geq \hbeta_t(\hat{\pi}(2)) \geq \cdots \geq \hbeta_t(\hat{\pi}(n))$.
We then apply this permutation to the nodes in the graph to obtain the sorted adjacency matrix $\hat{A_t}(i,j) = A_t(\hat{\pi}(i),\hat{\pi}(j))$, where $\bbA_t$ and $\hat{\bbA}_t$ represent the adjacency matrices of the original observed graph and the sorted version based on the underlying graphon, respectively.
We denote the set of sorted graphs by $\hat{\ccalD} = \{\hat{\bbA}_t\}_{t=1}^{M}$.

\paragraph{Step 2: Histogram approximation.}
After arranging the observed graphs based on the estimated latent variables, we aggregate the sorted adjacency matrices to obtain $M$ estimations of the underlying continuous graphon. 
To achieve this, we employ average pooling to create a histogram of the adjacency matrices following~\citet{sas}.
Specifically, denoting the histogram of the sorted adjacency matrix as $\hat{H}$, we have that
\begin{equation}\label{eq:hist}
\hat{H}(i,j) = \frac{1}{h^2} \sum_{s_1=1}^{h} \sum_{s_2=1}^{h} \hat{A}\left((i-1)h + s_1,(j-1)h + s_2\right)
\end{equation}
where $h$ is the size of the pooling window. 

When this operation is applied to a graph of size $n$, the resulting histogram $\hat{H}$ belongs to $\mathbb{R}^{k \times k}$, where $k = \lfloor \frac{n}{h} \rfloor$. 
This step leads us to a new dataset $\ccalI = \{\hat{H}_t\}_{t=1}^{M}$, which consists of all the histograms.
As a result of sorting all the observed graphs in the previous step, we can view the histogram dataset $\ccalI$ as a \emph{collection of discrete realizations (noisy sorted graphs) of the true graphon $\omega$, each captured at a different resolution}.
Essentially, we obtain discrete (noisy) observations of a continuous function at various points. 
Thus, we leverage INRs to obtain a continuous representation of the graphon $\omega$ in the following step. 

\paragraph{Step 3: INR training.}
In this last step, we use the dataset $\ccalI $ to learn $f_\theta(x,y): [0,1]^2 \rightarrow [0,1]$, the INR that represents the unknown graphon.
Recall that $\ccalI$ contains histograms of various sizes (resolutions), where each element within the matrices represents a 2-dimensional coordinate paired with its corresponding histogram value.
Since the graphon's domain is $[0,1]$, we normalize each coordinate by $k_t$, the total number of pixels in each dimension of $\hat{H}_t$.
These normalized coordinates, along with their corresponding values, form a supervised dataset of discrete points of the underlying continuous graphon. 
Each histogram provides $k_t^2$ data points, and the total number of data points across all histograms is $\sum_{t=1}^{M}k_t^2$.
We denote this dataset as $\ccalC$, where each data point is given by 
{blue}\begin{align*}
     \ccalC = \Large\{\left(\frac{i}{k_t}, \frac{j}{k_t}, \hat{H}_t(i,j)\right) :&\\
     &\hspace{-2cm}i,j \in \left\{1, \cdots, k_t\right\},  t \in \left\{1, \cdots, M\right\}\},
\end{align*}

We use $\ccalC$ to train our INR model, treating it as a regression problem.
Specifically, we minimize the mean squared error (MSE) between the predicted and true function values for each coordinate:
\begin{equation}\label{loss}
\begin{aligned}
    \theta^\star &= \argmin_{\theta} \; \sum_{t=1}^{M} w_t \sum_{i=1}^{k_t} \sum_{j=1}^{k_t} \left( \hat{H}_t(i,j) - f_\theta\left(\frac{i}{k_t}, \frac{j}{k_t}\right) \right)^2,
\end{aligned}
\end{equation}
Notice that for each graph $G_t$, we weigh the MSE by $w_t = \frac{|G_t|}{\sum_{i=1}^{M} |G_i|}$ to account for the effect of the resolution, where $|G_i|$ represents the number of nodes of the observed graph $i$. 
By weighting the MSE loss by the number of nodes, we force the model to increase the influence of the larger graphs, as their high resolution provides a better estimate of the true graphon.

\paragraph{SIGL for learning a parametric family of graphons.}
A similar training process can be used for learning a \emph{parametric} family of graphons \(\{ \omega_\alpha \}\) parameterized by some unknown parameter \(\alpha~\in~\mathbb{R}\).
The key difference lies in the fact that the observed graphs, originating from different graphons, exhibit distinct structural properties.
Thus, estimating a single graphon with an INR $f_\theta: [0,1]^{2}\rightarrow [0,1]$ cannot adequately capture the varying structures of the observed graphs.
To address this challenge, we modify the INR by considering a function that operates on higher-dimensional inputs, denoted as $f_{\theta}(x,y,\alpha): [0,1]^{3} \rightarrow [0,1]$, where $\alpha \in [0,1]$.
We learn the INR $f_{\theta}(x,y,\alpha)$ by following a procedure similar to the one described in Section~\ref{subsec:single_graphon}, but with some modifications:
$(i)$ instead of learning the latent variables directly for each $\alpha$, we use a pre-trained GNN for one particular $\alpha$ and exploit the GNN's transferability to generalize across other $\alpha$ values; and $(ii)$ we estimate the latent variable $z_t$ (a proxy of the true parameter $\alpha$ associated with a graph $G_t$) by computing the GW distance between graphs, enabling us to distinguish each $\omega_{\alpha}$.
We defer further details of the parametric case to Appendix~\ref{subsec:parametric_graphon}.

\noindent\textbf{Complexity analysis.} The first step of SIGL is sorting the observed graphs.
Being $L_1$ the number of epochs to train the GNN, and assuming $n_{T}$ the size of the largest graph in the training set, then the complexity is dominated by the largest graph, i.e., $\ccalO(L_1Mn_{T}^2)$.
The second step is the histogram approximation, which involves $M$ additions of $n_T^2$ evaluations: each bin's complexity is $\ccalO(h^2)$, and there are $k_T^2 = (\frac{n_T}{h})^2$ bins.
Thus, the complexity of step 2 is $\ccalO(Mn_T^2)$.
Finally, step 3 is the INR training, which is just an MSE loss function~\eqref{loss}.
Thus, the complexity is $\ccalO(L_2Mk_{T}^2)$, where $k_{T}$ is the resolution of the largest graph, which yields a final complexity $\ccalO(M(L_2k_{T}^2 + (L_1 + 1)n_T^2)$.
Assuming $L_1 = L_2 = L$, and a fixed-size pooling window $h$, it reduces to $\ccalO(MLn_{T}^2)$.
On the other hand, even using an efficient implementation of the GW distance~\citep{flamary2021pot}, the complexity of IGNR~\citep{ignr} is $\ccalO(ML(n_{T}^2k_{T} + k_{T}^2n_{T}))$\citep{peyre2016gromov}; assuming again a fixed $h$, the final complexity is $\ccalO(MLn_{T}^3) $.
A summary is described in Table~\ref{tab:comparison}.

\begin{table}[!htb]
    \centering
    \caption{\small{Comparison of our work with SAS, USVT, GWB, and IGNR. Our method combines the quadratic (and scalable) complexity of classical methods, and the resolution-free property due to the parametrization using an INR.}}
    \label{tab:comparison}
    \scalebox{0.85}{\begin{tabular}{c|c|l}
        \hline
        Method & Resolution-free & Complexity \\
        \hline
        SAS & \xmark & $\ccalO(n_T^2 + k_T^2\log k_T^2)$ \\
        USVT & \xmark & $\ccalO(n_T^3)$ \\
        GWB & \xmark & $\ccalO(ML(n_T^2 k_T + k_T^2n_T))$ \\
        IGNR & \cmark & $\ccalO(ML(n_T^2 k_T + k_T^2n_T))$ \\
        \textbf{SIGL (ours)} & \cmark & $\ccalO(ML(k_{T}^2 + n_T^2))$ \\
        \hline
    \end{tabular}}
\end{table}

\begin{figure*}[!htb]
    \centering
    \includegraphics[width=0.75\textwidth]{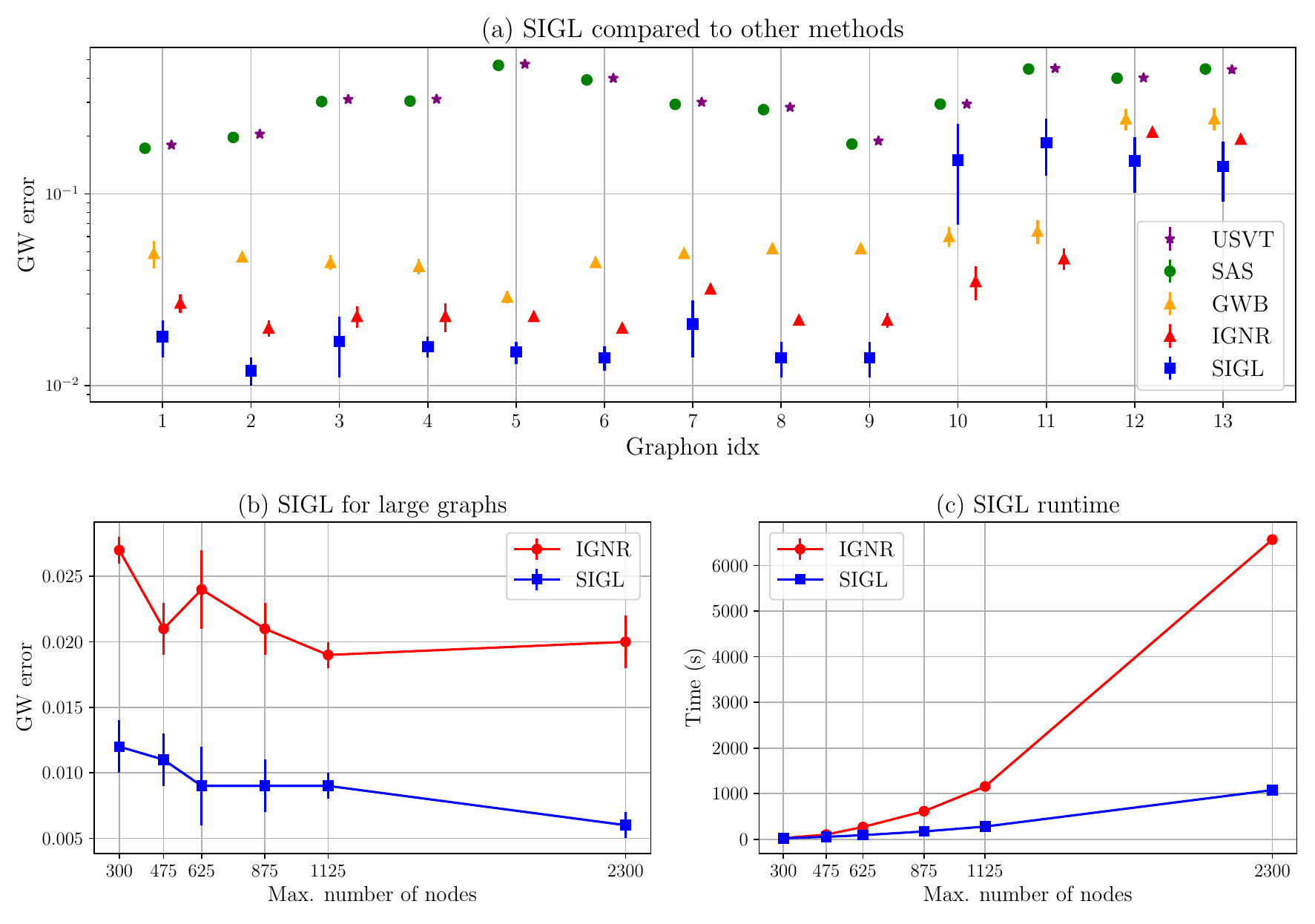}
    \caption{\small{Single graphon estimation.
    (a)~SIGL, compared to the other graphon estimation methods, demonstrates superior performance.
    (b)~Increasing the number of nodes helps to have a better estimation of the graphon. Also, our method still outperforms IGNR.
    (c)~SIGL estimates graphons faster than IGNR, especially in larger graphs.}}
    \label{fig:singlegraphon}
\end{figure*} 

\subsection{Asymptotic consistency of SIGL}
\label{subsec:theory}
\vspace{-0.05in}

In this section, we discuss the statistical consistency of SIGL.
As demonstrated later by experimental results, our method achieves better estimations and lower errors by increasing the number of nodes in the observed graph dataset. 
This behavior is characterized by the following proposition, which examines the framework's performance in the limit.

\begin{Proposition}[MSE of the estimator]\label{prop.1}
Let $\omega(x,y)\colon [0, 1]^2 \rightarrow [0,1]$ be the true graphon and $f_\theta(x,y)\colon [0, 1]^2 \rightarrow [0,1]$ be the estimated graphon by SIGL. Assuming that $\omega$ is Lipschitz continuous with constant $\ccalL>0$, and selecting the pooling window $h = \log n$ in Step 2, then,
\begin{align*}
    \lim_{n \rightarrow \infty} \mathbb{E} [\|f_\theta(x,y) - \omega(x,y) \|_2^2 ] \leq \varepsilon_{tr} + 4\ccalL^2\tau^2, 
\end{align*}
where $\varepsilon_{tr}$ is the maximum regression error of the INR on the training data points of $\ccalC$, $\tau$ is the maximum estimation error of GNN in Step 1 on the true latent variables of the nodes, and $n$ is the size of the largest graph in the dataset.
\end{Proposition}

Proposition \ref{prop.1} holds for estimating a single graphon based on multiple observed graphs. 
The expectation is taken over the sampled graphs used to build our estimate.
It is worth mentioning that multiple observed graphs provide various resolutions of the true graphon, aiding the GNN along with the auxiliary graphon in Step 1 to achieve lower $\tau$ and the INR in Step 3 in achieving a lower $\varepsilon_{tr}$. 
Since $n$ represents the size of the largest graph in the dataset, even with a single graph containing an infinite number of nodes, we can reach the limit mentioned above. 

Also, in this proposition, the model architecture influences the result through $\varepsilon_{tr}$ and $\tau$. 
More expressive INR and GNN architectures can yield lower values of $\varepsilon_{tr}$ and $\tau$, respectively, reducing the mean square error of the estimator.
This depends on the relationship between the true graphon $\omega(x,y)$ and the set of functions that the INR can express. 
If the true graphon is contained within or close to the expressive set of the INR, we can expect the training error $\varepsilon_{tr}$ to remain small even as the number of nodes grows arbitrarily. 
This proximity ensures that the INR can effectively approximate the true graphon, leading to a good estimator. 
Conversely, if there is a significant mismatch between the functions that the chosen INR can represent and the true graphon, the training error will be larger, resulting in a worse estimator. 
Our theoretical results make this relationship precise, highlighting the importance of selecting an appropriate model architecture for accurate graphon estimation.
Proofs and more details are provided in Appendix~\ref{sec:supp_proof}.

\vspace{-0.1in}
\section{EXPERIMENTS}
\label{sec:results}
\vspace{-0.1in}

In this section, we first compare SIGL with existing graphon learning methods for the task of estimating a single graphon in Section~\ref{subsec:exp_singlegraphon}.
Then, in Section~\ref{subsec:scalability}, we evaluate SIGL's performance on larger graphs to assess its scalability.
Finally, in Section~\ref{subsec:mixup}, we demonstrate the applicability of SIGL to real-world scenarios in the context of graph data augmentation.
We defer experiments on the estimation of a parametric family of graphons to Appendix~\ref{subsec:parametric_graphon}.

\vspace{-0.05in}
\subsection{Single graphon estimation}\label{subsec:exp_singlegraphon}
\vspace{-0.05in}

\begin{figure*}[!htb]
    \centering
    \includegraphics[width=0.75\textwidth]{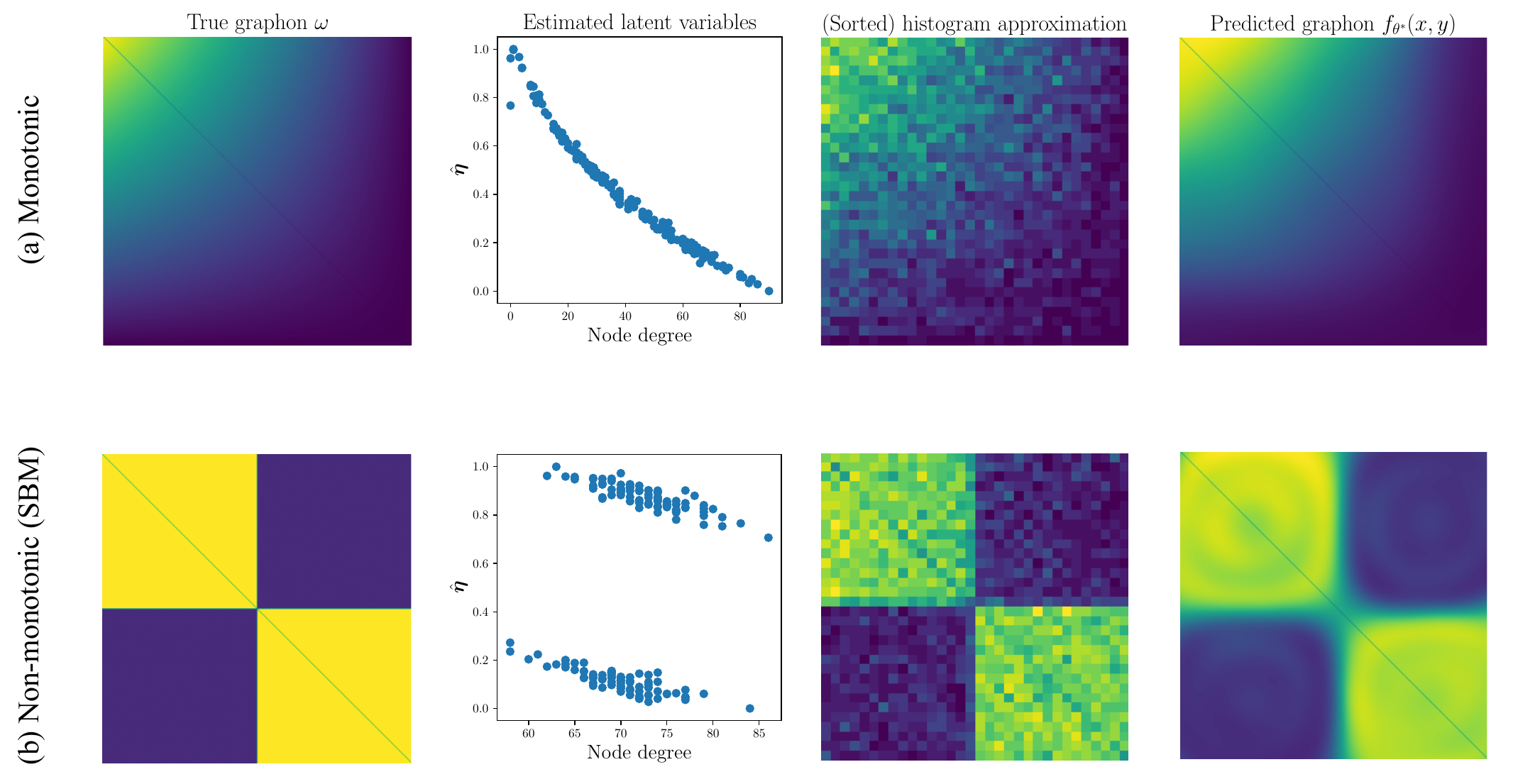}
    \caption{\small{Visualization of SIGL for: 
    (a)~a monotonic graphon, where the GNN learns latent variables that decrease as the node degree increases, and 
    (b)~a non-monotonic SBM graphon, where the GNN assigns similar latent variables to nodes within the same block, effectively seperating the block structure.
    }}
    \label{fig:estimatedgraphon}
\end{figure*}

\paragraph{Setup.}
To evaluate our model, we consider a set of 13 synthetic graphons (see Appendix~\ref{sec:supp1}), which are considered by~\cite{sas} and~\cite{ignr}.
Using these (ground-truth) graphons as generative models, we generate 10 graphs of varying sizes containing $\{75, 100, ..., 275, 300\}$ nodes from each graphon as described in Section~\ref{sec:background}.
Then, these graphs are fed to our pipeline as the input dataset to estimate the true graphon from which they are generated.
In estimating the histograms, we use $h= \log n$ as the size of the pooling window.
To evaluate this learned graphon, we sample both the true and estimated graphons with a resolution of 1000 on a grid in the unit square and calculate their distance by employing the GW distance~\citep{gw}.
To learn the latent variable, $g_{\phi_1}$ consists of three graph convolutional layers, followed by a ReLU activation function~\citep{gnnsurvey}. 
Additionally, the final layer is followed by a normalization layer to map the latent variables to the $[0,1]$ domain, similar to the true latent variables, $\eta$.
For the INR architecture, we employ SIREN~\citep{siren} with 2 layers, each comprising 20 neurons.
More details on the network structure and the training parameters are given in Appendix~\ref{sec:appendix_str}.

\vspace{-0.15in}
\paragraph{Baselines.}
We consider as baselines USVT~\citep{usvt}, SAS~\citep{sas}, IGNR~\citep{ignr} and GWB~\citep{gwb}.
To compare with IGNR, we sample the estimated graphons INR with a resolution of 1000. 
For SAS and USVT, the adjacency matrix of the observed graphs are zero-padded to the desired resolution (1000 in our case) before estimating the graphon.
In the case of GWB, the barycenter of the observed graphs is used to estimate the graphon, and we set the resolution to $K=\lfloor\frac{n}{\log n}\rfloor$, as suggested in~\citet{gwb}, and then we upsample the estimation to the desired resolution of 1000.
In all methods, we use the $GW$ distance between the estimated graphon and the true graphon as the estimation error.

\vspace{-0.1in}
\paragraph{Performance.}
As illustrated in Fig.~\ref{fig:singlegraphon}(a), SIGL achieves a lower GW distance between the estimated and the true graphon for the first nine graphons (1-9) and two of the last four graphons (10-13) compared to the baselines.

The first nine graphons (1-9) are monotonic, meaning the degree function follows a monotonic pattern when nodes are sorted by degree. 
Consequently, sorting by degree and minimizing the MSE~\eqref{loss} is expected to achieve a good performance.
Additionally, we expect that in Step 1 of the algorithm (see Section~\ref{subsec:single_graphon}), the latent variable estimation using $g_{\phi_1}$ will learn a latent variable that captures this monotonic ordering of the nodes, analogous to the degree. 
In contrast, the last four graphons (10-13) are non-monotonic.
As a consequence, all the nodes might have the same expected degree.
In these cases, \emph{while sorting by node degree will not work, the GNN in Step 1 is able to learn a meaningful sorting of the nodes}.
In particular, our approach enables us to distinguish nodes based on the graph structure, as seen in graphon 12 (an SBM).

We evaluate this behavior in Fig.~\ref{fig:estimatedgraphon} for graphons 1 and 12, representing monotonic and non-monotonic cases, respectively.
In the monotonic case, shown in Fig.~\ref{fig:estimatedgraphon}(a), the latent variables are learned in a way that follows the degree ordering of the nodes, with the latent variable decreasing monotonically as the degree increases.
In the SBM graphon, depicted in Fig.~\ref{fig:estimatedgraphon}(b), the GNN is able to distinguish nodes based on their block structure. 
Regardless of the degree, it assigns similar latent variables to nodes within the same block, enabling a good approximation using a histogram.

\begin{table*}[!htb]
    \centering
    \caption{\small{Performance comparisons of \textit{G}-Mixup with SIGL on different datasets. The metric is the classification accuracy.
    For REDD-M5, IGNR timed out after 5 hours; this is represented with $*$.}}
    \label{tab:gmixup}
    \begin{tabular}{lccccc}
        \hline
        \textbf{Dataset} & \textbf{IMDB-B} & \textbf{IMDB-M} & \textbf{{COLLAB}} & \textbf{REDD-B} & \textbf{REDD-M5} \\
        \hline
        \#graphs & 1000 & 1500 & 5000 & 2000 & 4999 \\
        \#classes & 2 & 3 & 3 & 2 & 5 \\
        \#avg.nodes & 19.77 & 13.00 & 74.50 & 429.63 & 508.52 \\
        \#avg.edges & 96.53 & 65.94 & 2457.22 & 497.75 & 594.87 \\
        \hline
        \multicolumn{6}{c}{\textbf{GIN}} \\
        \hline
        {vanilla} & 73.05{\small$\pm$2.54} & 49.02{\small$\pm$2.40} & 79.39{\small$\pm$1.24} & 91.78{\small$\pm$1.09} & 55.93{\small$\pm$0.89} \\
        {\textit{G}-Mixup {w/ USVT}} & 72.10{\small$\pm$3.83} & 47.89{\small$\pm$1.60} & 79.05{\small$\pm$1.25} & {91.32}{\small$\pm$1.51} & 55.74{\small$\pm$0.77} \\
        \textit{G}-Mixup {w/ IGNR} & 73.45{\small$\pm$3.07} & 49.10{\small$\pm$1.42} & 79.63{\small$\pm$1.51} & 91.35{\small$\pm$1.36} & * \\
        \textit{G}-Mixup w/ SIGL & \textbf{73.95}{\small$\pm$2.64} & \textbf{50.70}{\small$\pm$1.41} & \textbf{80.15}{\small$\pm$0.60} & \textbf{92.25}{\small$\pm$1.41} & \textbf{56.71}{\small$\pm$0.74} \\
        \hline
    \end{tabular}
\end{table*}

\vspace{-0.1in}
\subsection{Scalability.}
\label{subsec:scalability}
\vspace{-0.05in}
To evaluate the scalability of our method, we increase the number of nodes of the graphs in the input dataset.
This is a key factor in evaluating the applicability of our method to large-scale real-world graphs.
We generate 10 graphs with size of $\verb|offset|+\{75, 100, ..., 275, 300\}$, where $\verb|offset|$ is the ablation variable that modifies the number of nodes of all the generated graphs.
In particular, we consider $\verb|offset|=\{0, 175, 325, 575, 825, 2000\}$.
Our result in Fig.~\ref{fig:singlegraphon}(b) shows that with an increasing number of nodes, our method achieves a better estimation of the true graphon while still being superior to the second-best baseline in the previous section (IGNR). 
This outcome is consistent with Proposition~\ref{prop.1}, where we established that as the number of nodes increases, the estimation error is upper-bounded by the training error in the limit of infinite nodes. 
However, this enhancement in performance comes at the cost of increased computational time required for estimation. 
Therefore, it is important to analyze the behavior of 
SIGL's training time as the number of nodes in the input graphs increases.
We compute the average training time of SIGL for 10 input graphs over 10 trials and compare it with IGNR.
Note that all experiments were performed on the same computer.
As depicted in Fig.~\ref{fig:singlegraphon}(c), and supported by our complexity analysis in Section~\ref{subsec:single_graphon}, IGNR's training time rises faster (approximately cubic) due to its use of the GW distance in the loss function. 
In contrast, SIGL's complexity is quadratic in the number of nodes, which yields a lower training time for larger graphs and real-world applications.
Specifically, for graphs with over 2000 nodes, our method is approximately 6 times faster than IGNR.

\vspace{-0.12in}
\subsection{Real-world application: Graphon mixup}
\label{subsec:mixup}
\vspace{-0.05in}

Finally, to evaluate the performance of our method on real-world data, we consider graph data augmentation.
Mixup~\citep{zhang2017mixup, verma2019manifold} in the context of graphs has been applied with success in several works~\citep{gmixup, navarro2023graphmad}.
In a nutshell, the goal of graph mixup is to enhance the solution for downstream tasks by augmenting the dataset from existing data by following some strategy (more details in Appendix~\ref{app:mixup}); the more traditional is linear interpolation.
G-Mixup~\citep{gmixup} uses graphons to mix graphs from different classes, improving graph classification. 
Since graphs within each class share structural properties, learning a class-specific graphon is reasonable.
We replace graphon estimation in G-Mixup with our method and test it on four graph classification datasets.

For each dataset, we employ the GIN architecture as the GNN model to classify graphs. 
We consider three baselines: the vanilla case, where we use the original dataset without any augmentations; G-mixup with USVT, which corresponds to the original method from~\citet{gmixup}, and G-mixup with IGNR.
Using these learned graphons, we generate additional graphs with labels interpolated between the classes and train the model on this augmented dataset. 
More details on the training parameters are provided in Appendix~\ref{app:mixup}.
As shown in Table~\ref{tab:gmixup}, our method achieves improved performance on all of the datasets, demonstrating its effectiveness in learning representative graphons for each class.

One key advantage of SIGL compared to G-mixup with USVT is that, by using a resolution-free model for the graphon, we can generate graphs of arbitrary sizes to augment the training dataset. 
In contrast, G-mixup generates all augmented graphs at a fixed size because it relies on a fixed-resolution model of the graphon. 
While IGNR also models the graphon using an INR, which is resolution-free, the use of GW as the loss function makes it nearly impossible to learn a graphon in large real-world datasets like Reddit.
For example, in the Reddit datasets, it takes approximately 1 hour to approximate the graphon for each class, even with just 10\% of the dataset.
In contrast, SIGL requires about 3 minutes with the same number of graphs for each class.
Consequently, for the multi-class case of the Reddit dataset, IGNR computation timed out after 5 hours.

\vspace{-0.30in}
    \section{CONCLUSION}
\label{sec:Conclusion}
\vspace{-0.3in}
We introduced SIGL, a \emph{scalable} method for graphon learning using INRs as a resolution-free estimator.
SIGL leverages the histogram representation of sorted adjacency matrices as noisy measurements, combined with a GNN module learning an effective sorting.
We show that the proposed estimator is consistent with sufficiently powerful GNN and INR architectures. 
Extensive experiments on synthetic and real-world graphs demonstrate that SIGL outperforms state-of-the-art methods, achieving higher accuracy with reduced computational cost.

One limitation of our approach arises when the graph structures exhibit significant irregularity or heterogeneity.
In such cases, the sorting module struggles to accurately infer the latent variables, affecting the performance of SIGL.
Moving forward, a natural extension of our method involves considering cases where the graphs also have features defined at the nodes.
We can treat this scenario by jointly learning a graphon and the corresponding graphon signals, representing a key step toward enabling practical applications of graphon signal processing~\citep{xu2022signal, ruiz2021graphon}.

\section{ACKNOWLEDGMENTS}
This work was supported by the NSF under grants EF-2126387 and CCF-2340481.
NZ was partially supported by a Ken Kennedy Institute 2024–25 Ken Kennedy-HPE Cray Graduate Fellowship.

\bibliography{ref}

\section*{Checklist}

 \begin{enumerate}

 \item For all models and algorithms presented, check if you include:
 \begin{enumerate}
   \item A clear description of the mathematical setting, assumptions, algorithm, and/or model. Yes.
   \item An analysis of the properties and complexity (time, space, sample size) of any algorithm. Yes. In particular, for a complexity analysis, please refer to Section~\ref{sec:Methods}
   \item (Optional) Anonymized source code, with specification of all dependencies, including external libraries. Yes.
 \end{enumerate}

 \item For any theoretical claim, check if you include:
 \begin{enumerate}
   \item Statements of the full set of assumptions of all theoretical results. Yes.
   \item Complete proofs of all theoretical results. Yes. Proofs are provided in SM (Section~\ref{sec:supp_proof}).
   \item Clear explanations of any assumptions. Yes.   
 \end{enumerate}

 \item For all figures and tables that present empirical results, check if you include:
 \begin{enumerate}
   \item The code, data, and instructions needed to reproduce the main experimental results (either in the supplemental material or as a URL). Yes. Please refer to the source code if needed.
   \item All the training details (e.g., data splits, hyperparameters, how they were chosen). Yes. In particular in SM (Section~\ref{subsec:training}).
         \item A clear definition of the specific measure or statistics and error bars (e.g., with respect to the random seed after running experiments multiple times). Yes. Please refer to the source code if needed.
         \item A description of the computing infrastructure used. (e.g., type of GPUs, internal cluster, or cloud provider). Yes. In particular in SM (Section~\ref{subsec:training}).
 \end{enumerate}

 \item If you are using existing assets (e.g., code, data, models) or curating/releasing new assets, check if you include:
 \begin{enumerate}
   \item Citations of the creator If your work uses existing assets. Yes.
   \item The license information of the assets, if applicable. Yes.
   \item New assets either in the supplemental material or as a URL, if applicable. Yes.
   \item Information about consent from data providers/curators. Not Applicable
   \item Discussion of sensible content if applicable, e.g., personally identifiable information or offensive content. Not Applicable
 \end{enumerate}

 \item If you used crowdsourcing or conducted research with human subjects, check if you include:
 \begin{enumerate}
   \item The full text of instructions given to participants and screenshots. Not Applicable
   \item Descriptions of potential participant risks, with links to Institutional Review Board (IRB) approvals if applicable. Not Applicable
   \item The estimated hourly wage paid to participants and the total amount spent on participant compensation. Not Applicable
 \end{enumerate}

 \end{enumerate}

\vfill

\onecolumn
\aistatstitle{Scalable Implicit Graphon Learning \\
Supplementary Materials}

\begin{appendix}
\renewcommand{\theequation}{A.\arabic{equation}}
\setcounter{equation}{0}
\setcounter{problem}{1}

\vspace{-1cm}
\section{Implementation details.}
\label{sec:appendix_str}

\subsection{Table of True Graphons}
\label{sec:supp1}

\begin{table}[!htb]
\caption{Ground truth graphons.}
    \centering\small
    \begin{tabular}{lll}
        \toprule
         & $\omega(x,y)$ \\ 
         \midrule
1 &  $xy$  \\
2 &  $\exp(-(x^{0.7}+y^{0.7}))$  \\
3 &  $\frac{1}{4}(x^2+y^2+\sqrt{x}+\sqrt{y})$ \\
4 &  $\frac{1}{2}(x+y)$ \\ 
5 & $(1+\exp(-2(x^2+y^2)))^{-1}$\\ 
6 & $(1+\exp(-\max\{x,y\}^2-\min\{x,y\}^4))^{-1}$\\ 
7 & $\exp(-\max\{x,y\}^{0.75})$\\
8 & $\exp(-\frac{1}{2}(\min\{x,y\}+\sqrt{x}+\sqrt{y}))$\\
9 &  $\log(1+\max\{x,y\})$\\
\midrule
10 &  $|x-y|$\\
11 & $1-|x-y|$\\
12 & $0.8\mathbf{I}_2\otimes\mathds{1}_{[0,\frac{1}{2}]^2}$\\
13 & $0.8(1-\mathbf{I}_2)\otimes\mathds{1}_{[0,\frac{1}{2}]^2}$\\
        \bottomrule
    \end{tabular}
    \label{table:groundtruth}
\end{table}

\subsection{Network structure.}

Mathematically, our INR network is structured as

\begin{align*}
f_\theta(\mathbf{x}) = W_2(\phi_{1} \circ \phi_{0} )(\mathbf{x}) + b_2, \quad \phi_i(x_i) = \sin (W_ix_i + b_i).
\end{align*}

Here, $\mathbf{x} =(x,y) \in \mathbb{R}^{2}$ represents the coordinates of the pixels in the square unit. 
Additionally, $\phi_i: \mathbb{R}^{M_i} \rightarrow \mathbb{R}^{N_i}$ denotes the $i$-th layer of the network. 
This layer comprises the linear transform defined by the weight matrix $W_i \in \mathbb{R}^{N_i \times M_i}$ and the biases $b_i \in \mathbb{R}^{N_i}$ applied to the input $x_i \in \mathbb{R}^{M_i}$, followed by the sine nonlinearity applied to each component of the resulting vector.

The node representation update in the graph convolutional layers in step 1 can be mathematically defined as follows:
\[
\bbz_v^{(l+1)} = \mathrm{ReLU} \left(\left[\bbW_k \cdot {\text{Mean}}\left(\left\{
\bbz_u^{(l)}, {\forall} u\in \mathrm{Neigh}(v)\right\}\right), \bbB_k \bbz_v^{(l)}\right]\right), \quad {\forall} v\in \mathcal{V},
\]
where \( \bbz_v^{(l)} \) represents the node embedding of the node \( v \) at layer \( l \), \( \mathrm{Neigh}(v) \) represents the set of neighboring nodes of node \( v \), and \( \text{Mean} \) is an aggregation function that combines the embeddings of neighboring nodes.
Finally, to achieve the latent variable of each node, we normalize the output of the final layer as 
\[
\hbeta_i = \frac{\bbz_i^{(L+1)} - \min(\bbz^{(L+1)}) }{\max(\bbz^{(L+1)}) - \min(\bbz^{(L+1)})},
\]
where $L$ is the number of layers in GNN and $\bbz_i$ is the representation of node $i$.
To initialize the embedding $\mathbf{z}_i^{(0)}$, we sample it from a Gaussian distribution with zero mean and a standard deviation of 1, i.e., $\mathbf{z}_i^{(0)} \sim \mathcal{N}(0,1)$.

\subsection{Algorithm.}

We show the pseudocode of SIGL in Alg.~\ref{alg:sigl_training}.
As explained in Section~\ref{subsec:single_graphon}, our method involves three steps.
\begin{algorithm}[!htb]
	\caption{SIGL training}\label{alg:sigl_training}
	\begin{algorithmic}[1]
 	\Require $\ccalD = \{\bbA_t\}_{t=1}^M$\\
    \State $\verb|Step 1: Node sorting learning|$
    \State Sample $\bbX_t \sim \ccalN(0, \bbI)\;\; \forall t=1,\cdots, M$
    \For{$\ell = 1\; \text{to}\;  L_1$}
    \State $\ccalL_{step_1}(\phi) = \sum_{t=1}^{M} \frac{1}{n_t^2}\sum_{i,j=1}^{n_t} \left[ A_t(i,j) - h_{\phi_{2}}\left(\hat{\eta}_{i,t}(\phi_1),\hat{\eta}_{j,t}(\phi_1)\right)\right]^2$
    \State $\phi = \mathrm{OptimizerStep}(\ccalL_{step_1})$ 
    \EndFor
    \State Define node sorting as $\hbeta_t(\hat{\pi}(1)) \geq \hbeta_t(\hat{\pi}(2)) \geq \cdots \geq \hbeta_t(\hat{\pi}(n))$
    \State Sort matrices: $\hat{A_t}(i,j) = A_t(\hat{\pi}(i),\hat{\pi}(j))$\\
    \State $\verb|Step 2: Histograph approximation|$
    \For {$t = 1\; \text{to}\;  M$}
    \State $\hat{H}_t(i,j) = \frac{1}{h^2} \sum_{s_1=1}^{h} \sum_{s_2=1}^{h} \hat{A}_t(ih + s_1,jh + s_2)$
    \EndFor 
    \State Collect $\ccalI = \{\hat{H}_t\}_{t=1}^M$\\
    \State $\verb|Step 3: Graphon estimation|$
    \For{$\ell = 1\; \text{to}\;  L_2$}
    \State $\ccalL_{step_3}(\theta) = \sum_{t=1}^{M} w_t \sum_{i=1}^{k_t} \sum_{j=1}^{k_t} \left( \hat{H}_t(i,j) - f_\theta\left(\frac{i}{k_t}, \frac{j}{k_t}\right) \right)^2$
    \State $\theta = \mathrm{OptimizerStep}(\ccalL_{step_3})$
    \EndFor
	\Return $f_{\theta^*}(.,.)$
	\end{algorithmic}
\end{algorithm}

\subsection{Training settting.}
\label{subsec:training}
We use the $\verb|Adam|$ optimizer to train the model with a learning rate of $lr = 0.01$ for both Step 1 and Step 3, running for 100 epochs. 
In Step 1, the batch size consists of 1 graph, while in Step 3, we use 512 coordinates per batch.
In Step 1, the first component of the structure, $g_{\phi_1}$, consists of three consecutive graph convolutional layers~\citep{gcn}, each followed by a ReLU activation function. 
All convolutional layers have 16 hidden channels.
Both INR structures in Step 1 ($h_{\phi_2}$) and Step 3 ($f_{\theta}$) have 20 hidden units per layer, with a default frequency of 10 for the $\sin$ activation function.
Both INR structures in Step 1 ($h_{\phi_2}$) and Step 3 ($f_{\theta}$) have 20 hidden units per layer, with a default frequency of 10 for the $\sin$ activation function.
For each true graphon, we conduct 10 different trials of datasets, and the final result is the average across these trials.
The standard deviation across the trials is computed and represented as error bars in the result plots.
We train our model on a single GPU resource: NVIDIA GeForce RTX 2080 Ti with 12GB memory.

\subsection{Baselines methods.}

\begin{itemize}
    \item \textbf{SAS\citep{sas}.} 
    The primary hyperparameter in SAS is the size of the pooling window.
    We follow the original paper and set to $h = \log(n)$.

    \item \textbf{IGNR~\citep{ignr}.}
    We use the implementation from the original paper~\citep{ignr}.
    The INR structure is SIREN similar to our structure but with 3 layers of 20 units. 
    Also, the default frequency for $\sin$ activation is set to 30.

    \item \textbf{USVT\citep{usvt}.}
    Following~\citep{gmixup}, we set the threshold for the singular values to $0.2\sqrt{n}$, where $n$ represents the size of the largest graph.
    
\end{itemize}

\paragraph{Comparison between SIGL and SAS.} Steps 1 and 2 of our proposed method are similar to SAS. 
The main difference between ISGL and SAS is that SAS applies these steps to a single observed graph and estimates the true graphon as a discrete matrix of the same size as the observed graph.
In contrast, SIGL directly learns the latent variables to obtain a common ordering of the nodes across all observed graphs. 
Additionally, we apply these steps to a set of graphs, allowing us to approximate the true graphon at multiple resolutions and enabling the learning of a continuous representation.
The rationale behind these two steps is that as a graph increases in size, the sorted observed graphs should converge to the true graphon. 
Thus, we can estimate the true continuous graphon by sorting the given graphs and applying appropriate smoothing algorithms.

\paragraph{Implementation difference between SIGL and IGNR.}
Something important to remark between SIGL and IGNR is the practical implementation.
While our method is fully deployed in a GPU, IGNR needs to combine a GPU and CPU: the computation of the $GW$ distance has to be done in CPU (there is not an implementation in GPU).
This leads to an impractical method for real-world applications, as it is shown in g-mixup.

\section{Scalable Learning of a Parametric Family of Graphons}
\label{subsec:parametric_graphon}

In this section, we focus on a \emph{parametric} family of graphons \(\{ \omega_\alpha \}\) parameterized by some unknown parameter \(\alpha~\in~\mathbb{R}\), as described in Problem~\ref{prob2}.

\begin{problem}\label{prob2}
Given a set of graphs \(\ccalD = \{\bbA_t\}_{t=1}^{M}\) generated from a collection of graphons \(\{ \omega_\alpha \}\) parameterized by some unknown parameter \(\alpha \in \mathbb{R}\) but \emph{without} knowing which graph was generated by which graphon, estimate the collection \(\{ \omega_\alpha \}\).
\end{problem}

The key difference between Problem~\ref{prob2} and Problem~\ref{prob1} lies in the fact that the observed graphs, originating from different graphons, exhibit distinct structural properties.
Thus, estimating a single graphon with an INR $f_\theta: [0,1]^{2}\rightarrow [0,1]$ cannot adequately capture the varying structures of the observed graphs.
To address this challenge and solve Problem \ref{prob2}, we modify the INR by considering a function that operates on higher-dimensional inputs, denoted as $f_{\theta}(x,y,\alpha): [0,1]^{3} \rightarrow [0,1]$, where $\alpha \in [0,1]$.
We learn the INR $f_{\theta}(x,y,\alpha)$ by following a procedure similar to the one described in Section~\ref{subsec:single_graphon}, but with some modifications.

First, since the observed graphs are not generated from a common graphon, we cannot directly apply step 1.
Instead, we use an independent set of graphs generated from another \emph{single} graphon (chosen at random) in step 1 and minimize~\eqref{eq:loss0} over this set of independent graphs.
This gives us a pretrained GNN structure ($g_{\phi_1}$) along with a representation of the auxiliary graphon of those independent graphs ($h_{\phi_2}$).
This trained GNN shows transferability through experimental studies, meaning that it can be used to sort another set of graphs with similar structures generated from a different graphon. 
Further details on this transferability property are provided in Section~\ref{sec:pretrain_eval}.
As a result, we can use this $g_{\phi_1}$ to estimate the latent variable of the nodes and sort the graphs under study.
Then, we apply the same step 2 as described in Section~\ref{subsec:single_graphon}, obtaining the dataset $\ccalC$.

In the third step, we estimate the INR given the dataset $\ccalC$.
Compared to the single case, the INR depends on the parameter \(\alpha\): for each $\alpha$ we have a different graphon. 
The main challenge in this context arises from the lack of information about the parameter $\alpha_t$ associated with each given graph $G_t$.
Consequently, we first seek to capture the corresponding $\alpha_t$ by estimating a latent representation $z_t \in \mathbb{R}$.

\paragraph{Estimating the latent representation of $\alpha$.}
Given $\alpha$, each graphon $\omega_\alpha$ presents a distinct structure.
Thus, we need a method to serve as a proxy for distinguishing these differences and capturing $z_t$.
To achieve this, we use the auxiliary graphon \( h_{\phi_2} \) from Step 1 as a reference, measuring the deviation of each observed graph in the dataset from this reference.

Specifically, first we evaluate the auxiliary graphon \( h_{\phi_2} \) with resolution \( N \) denoted as $\bbB_r = h_{\phi_2}(s(N))$.
We consider regular-spaced coordinates in $[0,1]^2$, such that $s(N) = \{ \bbx_{i,j}=(x_i,y_j) | x_p=\frac{p-1}{N}, y_q=\frac{q-1}{N}, \forall p,q=1,\cdots,N \}$.
Then we calculate the $GW_2$ distance between the sampled graphon and each graph in our dataset \(d_t = GW_2(\bbA_t, \bbB_r)\), which is normalized to \([0,1]\) and used as the latent space representation for each graph \(z_t = \frac{d_t}{\sum_{k=1}^{M}d_k}\).
Importantly, the absolute value of this distance is not the primary focus. 
Instead, the normalized relative distance of different graphs in the dataset to the baseline helps us differentiate between them.
\emph{Note that since this value is computed only once for each graph in the dataset, it neither adds significant computational burden nor affects scalability.}

Once we calculated the latent representation, we use $f_\theta(x,y,z_t)$ to parameterize the graphon from which $\bbA_t$ is generated.
We learn the parameters $\theta$ by minimizing the following loss function:
\begin{align}
\label{eq:loss_parametric}
    \resizebox{0.48\textwidth}{!}{$
    \theta^\star = \argmin_{\theta} \; \sum_{t=1}^{M} w_t \sum_{i,j=1}^{k_t} \left( \hat{H}_t(i,j) - f_\theta\left(\frac{i}{k_t}, \frac{j}{k_t}, z_t \right) \right)^2.
    $}
\end{align}
Notice that the only difference between~\eqref{eq:loss_parametric} and~\eqref{loss} is the latent variable $z_t$. 
To summarize, the above training procedure works as follows:
$(i)$ Given a graph $\bbA_t$, we use a \emph{pretrained} GNN to obtain the latent variables $\hbeta$ of the nodes and sort the graph; 
$(ii)$ The $GW_2$ between $\bbA_t$ and the sampled auxiliary graphon (\( h_{\phi_2} \)) is computed, followed by a normalization to obtain $z_t$ for the given graph $t$;
$(iii)$ We build the training set ${\ccalC}$ which contains the histogram approximations of the graphs;   
$(iv)$ We train the INR by minimizing~\eqref{eq:loss_parametric}.

To evaluate our method in a parameterized setting, we consider $\omega_{\alpha}^{(1)}(x,y)=\exp \left(- \alpha (x^{0.7} + y^{0.7})\right)$, representing a parametric case of a monotonic graphon. 
Here, $\alpha$ is a scalar varying within the range $\{1, \frac{1}{1.2}, \frac{1}{1.4}, \frac{1}{1.6}, \frac{1}{1.8}, \frac{1}{2}\}$.
Additionally, we consider a graphon representing a stochastic block model (SBM) with two blocks, as a special case of non-monotonic graphons. 
This type of graphon can be defined as:
\begin{equation}
\begin{split}
    \omega_{\alpha}^{(2)}(x,y)= q \mathds{1}_{[0,1]^2}(x,y) + (p_1-q) \mathds{1}_{[0,\alpha]^2}(x,y)\\+(p_2-q)\mathds{1}_{[1-\alpha,1]^2}(x,y)
\end{split}
\end{equation}
where $\alpha$ is the size of the smaller block, and we use $p_1 = 0.8$, $p_2 = 0.8$, and $q = 0.1$.
We vary $\alpha$ within the range $\{0.1, 0.2, 0.3, 0.4\}$. 
Note that graphon 13 is the same as $\omega_{\alpha=0.5}^{(2)}$.

\paragraph{Dataset generation.}
For each value of $\alpha$, we conduct 10 trials, generating 10 graphs per trial with sizes ranging from 50 to 300 nodes.
All the generated graphs from different $\alpha$ values are combined to form a dataset comprising graphs originating from various graphons.
To pretrain the GNN model in step 1, we randomly select one graphon and repeat the single graphon case as described in the previous section. 
For $\omega_{\alpha}^{(1)}$, we randomly choose a graphon from the set $\{1, 3, 4, \cdots, 9\}$, excluding graphon 2 ($\omega_{\alpha=1}^{(1)}$). 
For the $\omega_{\alpha}^{(2)}$, we use graphon 12 as the pretraining graphon.

\subsection{Evaluation.}\label{sec:pretrain_eval}

To evaluate the parameterized scenario, we combine all the generated graphs from various $\alpha$ values and trials, then split this combined dataset into training and test datasets with a ratio of 0.2 for the test dataset. 
After training the INR structure in step 2 using the training dataset, we estimate the true graphon for each test graph based on its latent representation.
Then, we group the testing graphs based on their true $\alpha$ value, averaging the error for each $\alpha$. 
On the other hand, as shown in the previous section, our method exhibited strong performance in the single graphon scenario. 
As a result, we partition the dataset for each $\alpha$ in the test data, assessing the estimated graphon for that specific $\alpha$ individually using our single graphon estimation method.
This enables a comparison between the single graphon and parameterized cases for each $\alpha$.

\begin{figure*}[!th]
    \centering
    \includegraphics[width=0.9\textwidth]{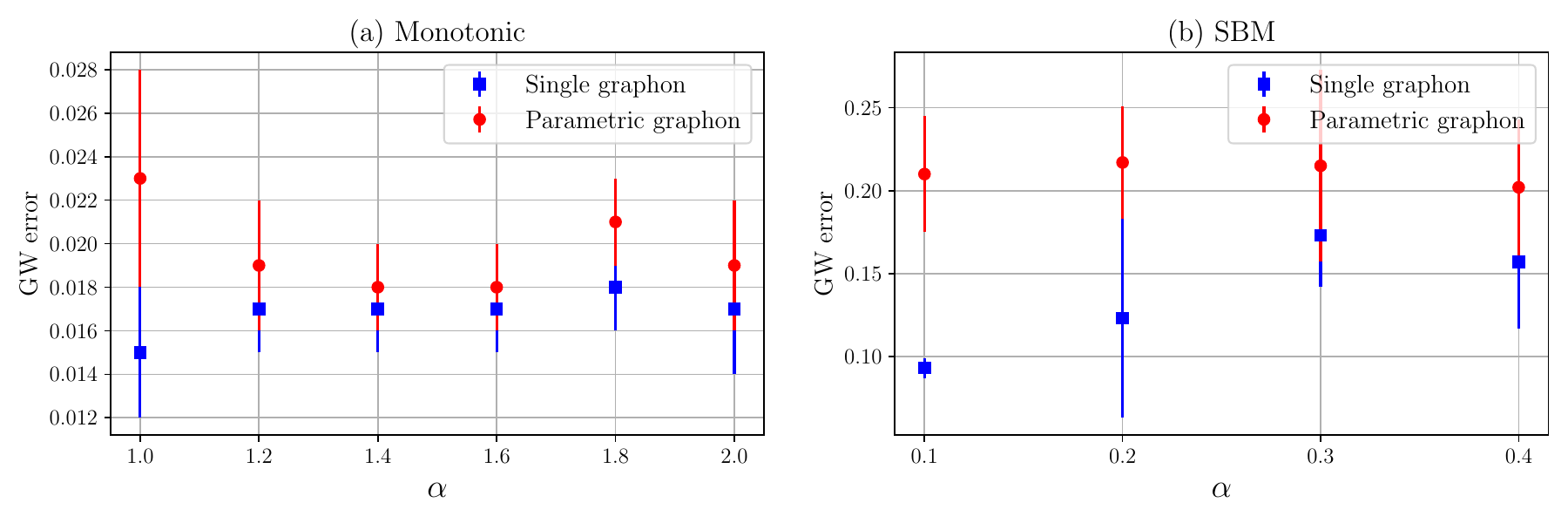}
    \caption{SIGL performs similarly to single graphon estimation when estimating the graphon in the parameterized scenario.
    (a)~Parametrized monotnic graphon.
    (b)~Parametrized SBM (non-monotonic) graphon.}
    \label{fig:paramgraphon}
\end{figure*}

\begin{figure}[!th]
    \centering
    \includegraphics[width=0.8\textwidth]{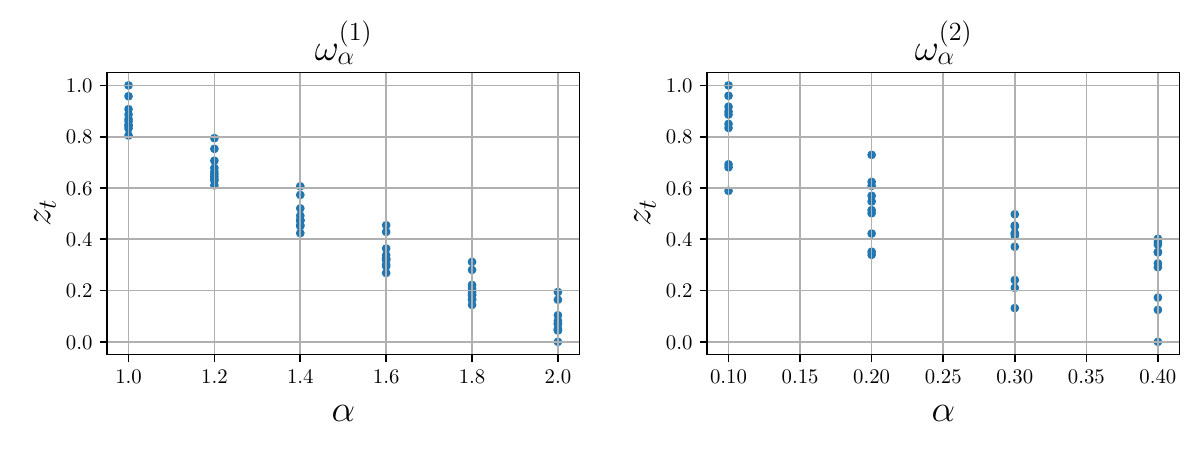}
    \caption{Latent space for the graphs are distinct for different true values of $\alpha$.}
    \label{fig:latentspace_alpha}
\end{figure}

As shown in Figure\ref{fig:paramgraphon}, for all $\alpha$ values, our method achieves an error in estimating the true graphon very close to that of the single graphon case. 
This demonstrates the efficacy of our method in distinguishing between graphs generated from different sources and in learning high-dimensional graphons for parameterized scenarios.
Note that for the monotonic case, we are plotting the denominator of $\alpha$ on the x-axis.
Additionally, we calculate $z$ for each observed graph and plot this latent variable against the true values of $\alpha$.
The results are shown in Fig.~\ref{fig:latentspace_alpha}, where it is evident that our model captures distinct latent spaces for graphs generated from different sources.

\begin{figure*}[!th]
    \centering
    \includegraphics[width=0.8\textwidth]{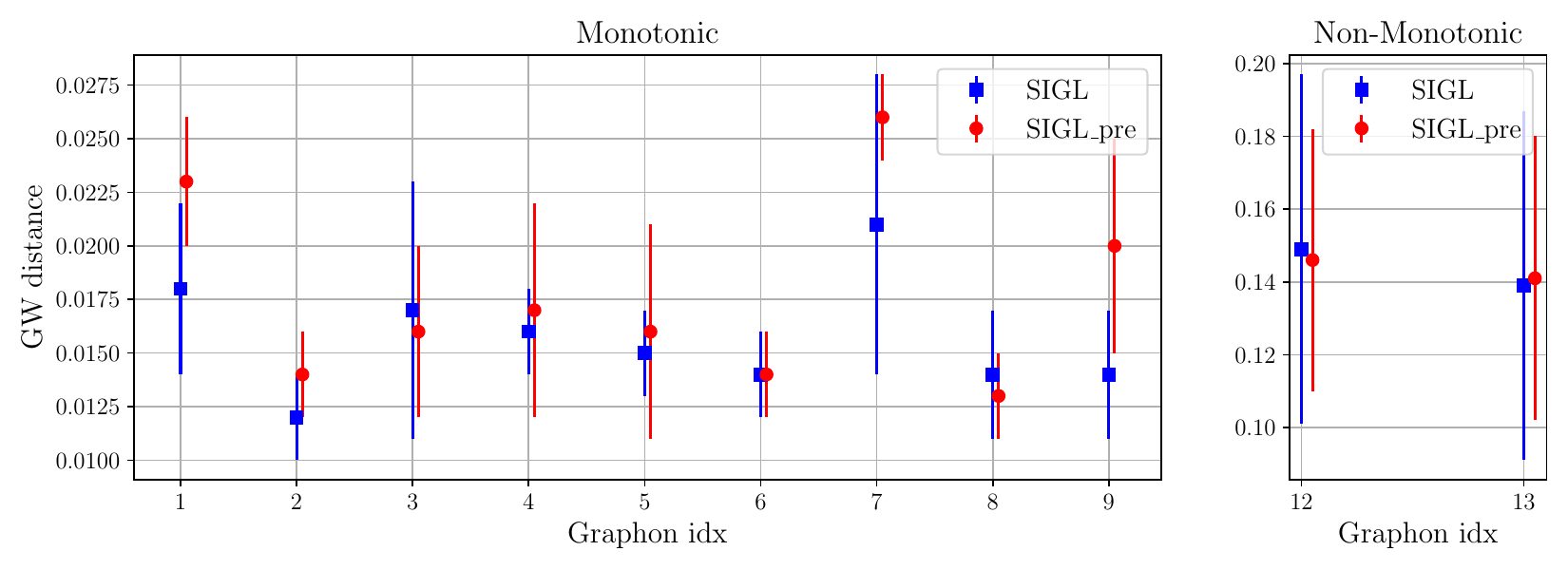}
    \caption{Transferability of the latent variable estimation step.}
    \label{fig:transferability}
\end{figure*}

\paragraph{Transferability of GNN} 
We study the transferability property of the GNN in Step 1 by conducting an experiment similar to that in Section~\ref{subsec:exp_singlegraphon}.
The key difference is that in Step 1, we randomly select a different synthetic graphon, generate observed graphs from it, and train the GNN-INR structure using these independent graphs. 
We then use the pretrained GNN to obtain the latent variables of the graphs under study. 
Steps 2 and 3 are performed in the same way as before.
We compare the graphon estimation performance compared to the main method.
The results are shown in Fig.~\ref{fig:transferability}.
We observe that using pretraining allows us to estimate the graphons with results that closely match those of the main procedure, highlighting the transferability of Step 1.

\section{Additional experiments}
\label{sec:supp_plots}

\subsection{Ablation analysis} 

\subsubsection{Performance when using less graphs}
\begin{figure}[!th]
    \centering
    \includegraphics[width=0.5\textwidth]{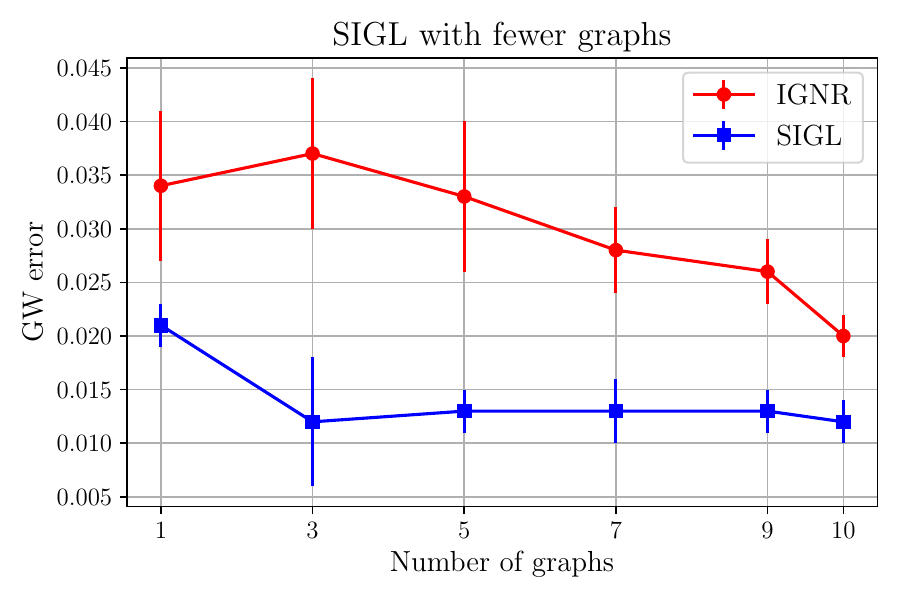}
    \caption{Effect of number of graphs in the estimation error}
    \label{fig:singlegraphon_2}
\end{figure}

We analyze the behavior of ISGL when having less number of graphs for the graphon estimation. 
The result is shown in Fig.~\ref{fig:singlegraphon_2} for graphon 1 from Table~\ref{table:groundtruth}.
We observe that even with as few as 3 graphs in the dataset, ISGL performs similarly to the case with 10 graphs. 
In contrast, since IGNR trains its structure based on the GW distance between observed and generated graphs, it requires a sufficient number of graphs to capture the graphon representation accurately, and with fewer graphs, its performance is limited.

\subsubsection{Changing the pooling factor for histogram approximation}
\begin{figure}[!th]
    \centering
    \includegraphics[width=0.5\textwidth]{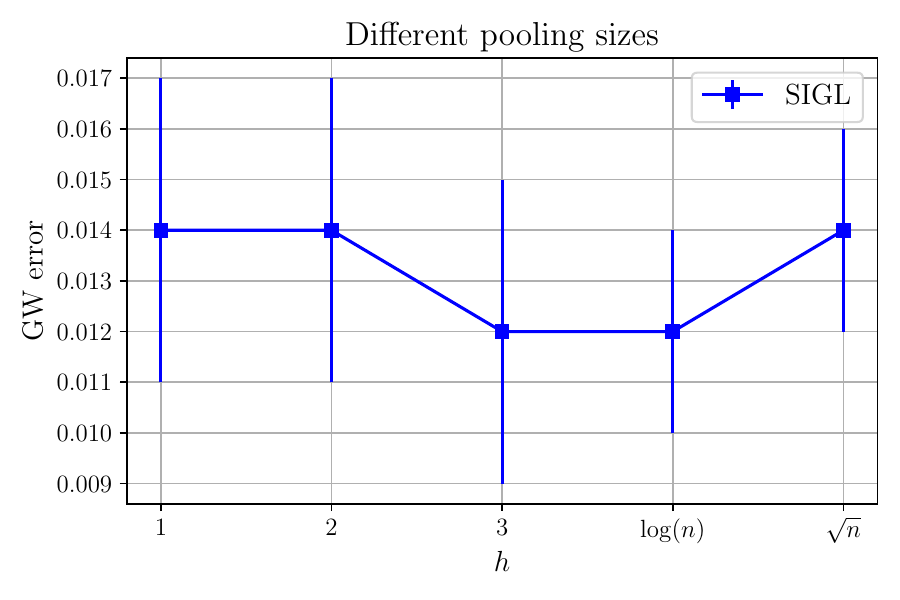}
    \caption{Impact of various pooling window sizes on the performance of the SIGL.}
    \label{fig:hsize}
\end{figure}

In this ablation study, we analyze the performance of SIGL when changing the pooling factor of the histogram.
The result is shown in Fig.~\ref{fig:hsize} for graphon 1 from Table~\ref{table:groundtruth}.
We observe that our method is robust to the size of the window with similar results for different choices for $h$.

\subsection{Additional results}

\begin{figure}[!th]
    \centering
    \includegraphics[width=0.8\textwidth]{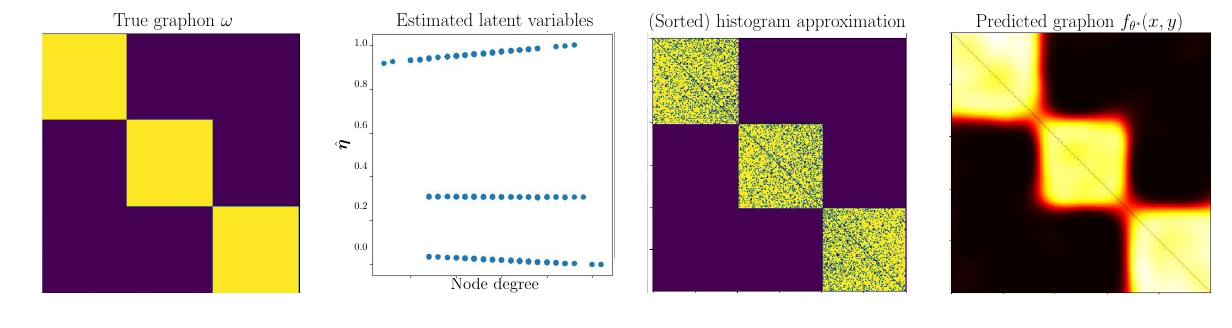}
    \caption{SIGL performance on SBM with 3 communities.}
    \label{fig:sbm3}
\end{figure}

Here, we consider another case involving non-monotonic graphons. 
Specifically, we analyze a graphon representing a stochastic block model (SBM) with three blocks and replicate results similar to those in Figure~\ref{fig:estimatedgraphon}.
As shown in Figure~\ref{fig:sbm3}, SIGL effectively learns meaningful latent variables for the nodes across different blocks, even when they share the same degree. 
Consequently, we achieve an estimation that closely approximates the true graphon.

\begin{figure}[!th]
    \centering
    \includegraphics[width=0.8\textwidth]{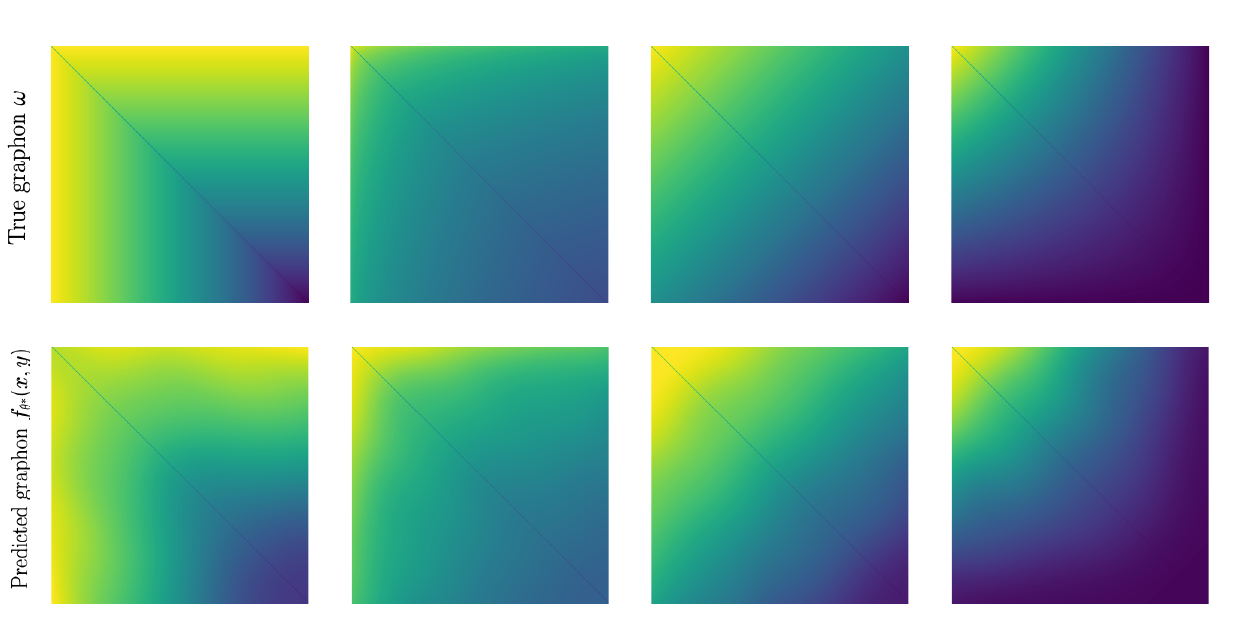}
    \caption{Comparison between the estimated and true graphons on four different samples of Table~\ref{table:groundtruth}.}
    \label{fig:moregraphon}
\end{figure}

Moreover, in Figure~\ref{fig:moregraphon}, we present four additional monotonic graphons along with their corresponding estimations using SIGL.

\subsection{Failure cases}

\begin{figure}[!th]
    \centering
    \includegraphics[width=0.5\textwidth]{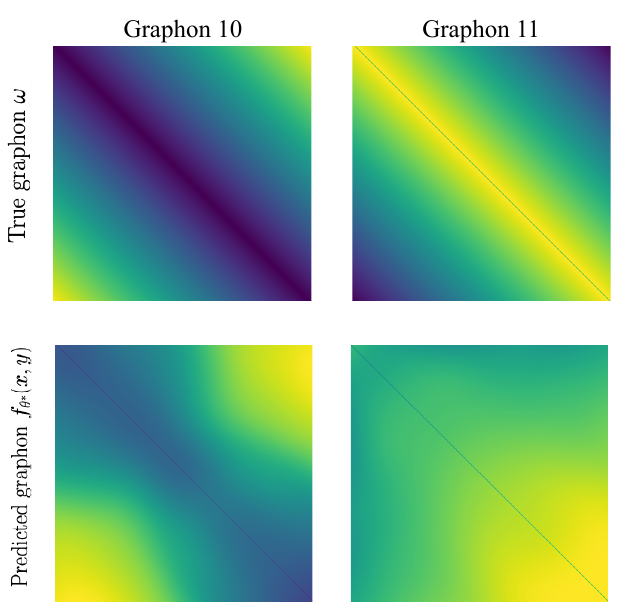}
    \caption{SIGL performance for graphons 10 and 11 from Table~\ref{table:groundtruth}.}
    \label{fig:Failure}
\end{figure}

As shown in Figure~\ref{fig:singlegraphon}, SIGL encounters difficulties with graphons 10 and 11. 
These two graphons, along with SIGL's estimations, are depicted in Figure~\ref{fig:Failure}.
This challenge may stem from the difficulty in sorting the graphs generated by these graphons. 

\section{G-mixup}
\label{app:mixup}

Training modern neural networks demands a large volume of data.
While in some cases there are enough data to successfully train a model, in several scenarios, we need to create more training data to improve the generalization of the trained model.
In this endeavor, data augmentation akin to Mixup aims to generate new training data from existing ones by following different strategies.
In summary, the more common mixup strategy creates new synthetic training samples through linear interpolation between pairs of randomly chosen data points. 
Formally, Mixup can be represented as follows:
\begin{equation}
    x_{\text {new }}=\lambda x_i+(1-\lambda) x_j, \quad y_{\text {new }}=\lambda y_i+(1-\lambda) y_j
\end{equation}
where $(x_i,y_i)$ and $(x_j,y_j)$ are two randomly selected samples with corresponding one-hot encoded labels.
This has been applied in several \emph{Euclidean} domains, ranging from images to natural language processing.
In the context of graphs, the nature of graph data poses significant challenges: 1) it is inherently irregular (non-Euclidean), where different graphs often have varying numbers of nodes; 2) graphs are not naturally aligned, making it difficult to correspond nodes between graphs; and 3) the topology of graphs from different classes can vary widely, while graphs within the same class often share similar structures.
To overcome these challenges, G-Mixup proposes to perform graph data augmentation by mixing graphons, which can be summarized as follows:
\begin{align}
\text { Graphon Estimation: } & \mathcal{G} \rightarrow W_{\mathcal{G}}, \mathcal{H} \rightarrow W_{\mathcal{H}} \\
\text { Graphon Mixup: } & W_{\mathcal{I}}=\lambda W_{\mathcal{G}}+(1-\lambda) W_{\mathcal{H}} \\
\text { Graph Generation: } & \left\{I_1, I_2, \cdots, I_m\right\} \stackrel{\text { i.i.d }}{\sim} \mathbb{G}\left(K, W_{\mathcal{I}}\right) \\
\text { Label Mixup: } & \mathbf{y}_{\mathcal{I}}=\lambda{\mathbf{y}_{\mathcal{G}}}+(1-\lambda) \mathbf{y}_{\mathcal{H}}
\end{align}
While the original work leverages fixed-resolution graphons, in this work, we use our resolution-free estimator.

\paragraph{Training parameters}
To ensure a fair comparison, we use the same hyperparameters for model training and the same architecture across vanilla models and other baselines. 
Also, we conduct the experiments using the same hyperparameter values as in~\citet{gmixup}.
For graph classification tasks, we employ the \verb|Adam| optimizer with an initial learning rate of 0.01, which is halved every 100 epochs over a total of 800 epochs. 
The batch size is set to 128. The dataset is split into training, validation, and test sets in a 7:1:2 ratio. 
The best test epoch is selected based on validation performance, and test accuracy is reported over eight runs with the same $\verb|seed|$ used in~\citet{gmixup}.

In G-Mixup, we generate 20\% more graphs for training. 
The graphons are estimated from the training graphs, and we use different $\lambda$ values in the range [0.1, 0.2] to control the strength of mixing in the generated synthetic graphs.
For graphon estimation using SIGL and IGNR, we use 20\% of the training data per class to estimate the graphon. 
The new graphs are generated with the average number of nodes as defined for the primary G-Mixup case, which is identified as the optimal size. 
However, for SIGL and IGNR, as the graphon is represented by a continuous function (INR) we generate graphs with sizes varying from the average number of nodes to the maximum number of nodes.

\section{Consistency Analysis}
\label{sec:supp_proof}

\subsection{Definitions}
\begin{itemize}
    \item $\omega: [0,1]^2 \rightarrow [0,1] $: True unknown graphon.

    \item $\omega_r \in \mathbb{R}^{r \times r}$: Discretization of the true graphon with resolution $r$. 

    \begin{align*}
        \omega_r(i,j) = \omega(\frac{i}{r}, \frac{j}{r}) \quad \forall i,j=1:r
    \end{align*}
    
    \item $\ccalD = \{A_t\}_{t=1}^{M}$: Set of adjacency matrix of the observed graphs generated from $\omega$ with sizes $(n_1, n_2, \cdots, n_M)$. 
    
    \item \textbf{Note:} In all of the following definitions, we assume that $n = \max(n_1, n_2, \cdots, n_M)$, which is the size of the largest graph in the set $\ccalD$. Also, all of the definitions are based on the largest graph of the observed graphs.

    \item $f_\theta(x,y): [0,1]^2 \rightarrow [0,1]$: Estimated graphon using SIGL, represented by an INR. 
    
    \item $\hat{\omega_r}^{est} \in \mathbb{R}^{r \times r}, \hat{\omega_k}^{est} \in \mathbb{R}^{k \times k}$: Discretization of the estimated graphon, $f_\theta(x,y)$, with resolutions $r$ and $k$, respectively. 
    
    \item $\hat{\pi}$: Permutation based on the estimated latent variable $\hbeta$, such that $\hbeta(\hat{\pi}(1)) \geq \hbeta(\hat{\pi}(2)) \geq \cdots \geq \hbeta(\hat{\pi}(n))$.
    
    \item $\hat{A} \in \mathbb{R}^{n \times n}$: Sorted adjacency matrix based on $\hat{\pi}$.
    $\hat{A}(i,j) = A(\hat{\pi}(i), \hat{\pi}(j))$.
    
    \item $\hat{H} \in \mathbb{R}^{k \times k}$: Histogram approximation of the sorted adjacency matrix based on degree permutation defined as in~\eqref{eq:hist}.

    \item $\pi$: Oracle permutation, such that $\bbeta(\pi(1)) < \bbeta(\pi(2)) < \cdots < \bbeta(\pi(n))$, where $\bbeta$ is the true latent variable of the nodes.
    
    \item $\tilde{A} \in \mathbb{R}^{n \times n}$: Sorted adjacency matrix based on the oracle permutation $\pi$.
    $\tilde{A}(i,j) = A({\pi}(i),{\pi}(j))$.
    
    \item ${H} \in \mathbb{R}^{k \times k}$: Histogram approximation of the sorted adjacency matrix based on oracle permutation similar to~\eqref{eq:hist} using $\tilde{A}$ instead of $\hat{A}$.
    \begin{align}\label{H}
        {H}_{ij} &= \frac{1}{h^2}\sum_{i_1=1}^{h}\sum_{j_1=1}^{h} {\tilde{A}}(ih+i_1, jh + j_1)
    \end{align}
    where similarly $h$ is the block size and $k = \frac{n}{h}$.
    
    \item $H^{\omega_r} \in \mathbb{R}^{k \times k}$: Histogram approximation of the sampled ground-truth graphon ($\omega_r$). 
    \begin{align}\label{H_omega}
        {H}^{\omega_r}(i,j) &= \frac{1}{h'^2}\sum_{i_1=1}^{h'}\sum_{j_1=1}^{h'} {\omega}_{r}(ih'+i_1, jh'+j_1)
    \end{align}
    where $h' = \frac{r}{k}$, which results in ${h' = \frac{r}{n} h}$.
    
\end{itemize}

\subsection{Background}

\begin{lemma}[Piecewise Constant Function Approximation]\label{lemma1}\citep[Lemma 1]{sas}
Let $\omega_r \in [0, 1]^{r \times r}$ be the sampled true graphon with resolution $r$, and let $H^{\omega_r} \in [0, 1]^{k \times k}$ be the histogram approximation defined in~\eqref{H_omega}. Then,
\begin{equation}
    \lVert H^{\omega_r} \otimes \mathbf{1}_{h' \times h'} - \omega_r \rVert_{2}^{2} \leq \frac{C'}{k^2},
\end{equation}
where $C'$ is a constant independent of $n$.
\end{lemma}

\begin{lemma}[Bounds on $\lVert H - H^{\omega_r} \rVert_2^2$]\label{lemma2}\citep[Lemma 4]{sas}
Let $H^{\omega_r}$ be the step function approximation of the sampled true graphon $\omega_r$, and $H$ be the histogram defined as~\eqref{H} using $A$. Then,
\begin{equation}
    \mathbb{E}\left[\lVert H - H^{\omega_r} \rVert_2^2\right] \leq \frac{k^4}{n^2}.
\end{equation}
\end{lemma}

\begin{lemma}[Bounds on $\lVert \hat{H} - H \rVert_2^2$]\label{lemma3}
Assume that the true graphon $\omega$ is Lipschitz continuous with a constant $\ccalL>0$. If $H$ and $\hat{H}$ are defined according to ~\eqref{H} and~\eqref{eq:hist} using $\tilde{A}$ and $\hat{A}$ respectively, then
\begin{equation}
    \mathbb{E}\left[\lVert \hat{H} - H \rVert_2^2\right] \leq
    \frac{k^4}{n^2} \left(2 + 4\ccalL^2\tau^2\right) + \left(4k^2\ccalL^2\tau^2\right) 
\end{equation}
where $\tau$ is the maximum ordering error of the nodes compared on the true latent variables.
\end{lemma}

\textit{Proof.}
By definition of $\hat{H}$ and $H$ we have:
\begin{equation}\label{HhatH}
    \mathbb{E}\left[\lVert \hat{H} - H \rVert_2^2\right] = \mathbb{E}\left[\sum_{i,j=1}^{k}\left(\frac{1}{h^2}\sum_{i_1,j_1=1}^{h}\left(\hat{A}(ih+i_1, jh + j_1)-\tilde{A}(ih+i_1, jh + j_1)\right)\right)\right]
\end{equation}

To evaluate the above expression, we need to estimate $\mathbb{E}\left[\left(\hat{A}(i, j)-\tilde{A}(i, j)\right)^2\right]$ and $\mathbb{E}\left[\hat{A}(i, j)-\tilde{A}(i,j)\right]$ for all $i,j=1,2,...,k$.
Let $\omega$ be the true graphon and
\begin{align}
    \hat{\omega}(i,j) = \omega\left(\bbeta(\hat{\pi}(i)),\bbeta(\hat{\pi}(j))\right)
\end{align}
be the graphon ordered by the estimated permutation of $\hat{\pi}$ in step 1 using ${\hbeta}$. Then it holds that:
\begin{align*}
    \mathbb{E}\left[\left(\hat{A}(i, j)-\tilde{A}(i, j)\right)^2\right] = \mathbb{E}\left[\left(\hat{A}(i, j)- \hat{\omega}(i,j)\right)^2 + \left(\tilde{A}(i, j)- {\omega}(i,j)\right)^2 + \left(\hat{\omega}(i, j)- {\omega}(i,j)\right)^2 \right]
\end{align*}
because $\mathbb{E}\left[\hat{A}(i, j)\right]=\hat{\omega}(i,j)$ and $\mathbb{E}\left[\tilde{A}(i, j)\right]={\omega}(i,j)$.

To bound the upper expression, we have:
\begin{align*}
    \mathbb{E}\left[\left(\hat{A}(i, j)- \hat{\omega}(i,j)\right)^2 \right] = Var\left[\hat{A}(i,j)\right] \leq 1, \\
    \mathbb{E}\left[\left(\tilde{A}(i, j)- {\omega}(i,j)\right)^2 \right] = Var\left[\tilde{A}(i,j)\right] \leq 1. \\
\end{align*}
Next, we bound $\left(\hat{\omega}(i, j)- {\omega}(i,j)\right)^2$ as:
\begin{align*}
    \left(\hat{\omega}(i, j)- {\omega}(i,j)\right)^2 &= \left[\omega\left(\bbeta(\hat{\pi}(i)),\bbeta(\hat{\pi}(j))\right) - \omega\left(\bbeta(i),\bbeta(j)\right) \right]^2 \\
    &\leq \left[\ccalL\left(|\bbeta(\hat{\pi}(i))-\bbeta(i)|+|\bbeta(\hat{\pi}(j))-\bbeta(j)|\right)\right]^2 
\end{align*}
where the inequality holds because of the Lipschitz condition of $\omega$.
Now, if we define $\tau$ as follows:
\begin{align*}
    \tau &= \max_i |\bbeta(\hat{\pi}(i))-\bbeta(i)|
\end{align*}
which is essentially the maximum of the estimation error of the true latent variables learned by GNN is step 1; then we have:
\begin{align*}
    \left(\hat{\omega}(i, j)- {\omega}(i,j)\right)^2 \leq 4\ccalL^2\tau^2
\end{align*}

Based on the above expressions, we have:
\begin{align*}
    \mathbb{E}\left[\left(\hat{A}(i, j)-\tilde{A}(i, j)\right)^2\right] \leq 2 + 4\ccalL^2\tau^2
\end{align*}
Similarly, $\mathbb{E}\left[\hat{A}(i, j)-\tilde{A}(i,j)\right]$ can be bounded as:
\begin{align*}
    \mathbb{E}\left[\hat{A}(i, j)-\tilde{A}(i, j)\right] = \mathbb{E}\left[\left(\hat{A}(i, j)- \hat{\omega}(i,j)\right) + \left(\tilde{A}(i, j)- {\omega}(i,j)\right) + \left(\hat{\omega}(i, j)- {\omega}(i,j)\right) \right] \leq 2L\tau.
\end{align*}
Now, going back to~\ref{HhatH}, we can show that:
\begin{align*}
    &\mathbb{E}\left[\sum_{i,j=1}^{k}\left(\frac{1}{h^2}\sum_{i_1,j_1=1}^{h}\left(\hat{A}(ih+i_1, jh + j_1)-\tilde{A}(ih+i_1, jh + j_1)\right)\right)\right] \\
    &\leq \sum_{i,j=1}^{k} \frac{1}{h^4} \left(h^2 \left(2+4\ccalL^2\tau^2\right)+ \frac{h^2(h^2-1)}{2} \left(2L\tau\right)^2\right) \\
    &\leq \frac{k^2}{h^2}\left(2+4\ccalL^2\tau^2\right) + k^2 \left(4\ccalL^2\tau^2\right).
\end{align*}
Substituting $n=kh$, we have:
\begin{align*}
    \mathbb{E}\left[\lVert \hat{H} - H \rVert_2^2\right] \leq
    \frac{k^4}{n^2} \left(2 + 4\ccalL^2\tau^2\right) + \left(4k^2\ccalL^2\tau^2\right) 
\end{align*}

\begin{lemma}[Bounds on $\lVert \hat{\omega}^{est}_k - \hat{H} \rVert_2^2$]\label{lemma4}
Let $\hat{\omega}^{est}_k$ be sampled from the estimated graphon, $f(x,y)$, with resolutions $k$ and $\hat{H}$ be the histogram approximation of sorted graphs defined as~~\eqref{eq:hist}. Then,

\begin{align*}
    \lVert \hat{\omega}^{est}_k - H^{\omega_r} \rVert_2^2 &\leq k^2 \varepsilon_{tr}, 
\end{align*}    
where $\varepsilon_{tr}$ is the maximum estimation error of INR on the training data points of $\hat{H}$.
\end{lemma}

\textit{Proof.}
\begin{align*}
    \lVert \hat{\omega}^{est}_k - \hat{H} \rVert_2^2 = \sum_{i=1}^{k}\sum_{j=1}^{k} |f((i-\frac{1}{2})h,(j-\frac{1}{2})h) - \hat{H}(i,j)|^2
    &\leq k^2 \varepsilon_{tr} \\
\end{align*}
where $\varepsilon_{tr}$ is defined as follows:
\begin{align*}
    \varepsilon_{tr} = \max_{i,j} |f_\theta((i-\frac{1}{2})h,(j-\frac{1}{2})h) - \hat{H}(i,j)| 
\end{align*}
essentially representing the maximum estimation error on the data points on which the INR is trained.


\begin{lemma}[Bounds on different resolutions of the estimated graphon]\label{lemma5}
Let $\hat{\omega}^{est}_{r}$ and $\hat{\omega}^{est}_k$ be the discretization of the estimated graphon, $f(x,y)$, with resolutions $r$ and $k$, respectively, where $f(x,y)$ is $\tilde{\ccalL}$-Lipschitz. Then,

\begin{align*}
    \lVert \hat{\omega}^{est}_{r} -  \hat{\omega}^{est}_k \otimes  \mathbf{1}_{h' \times h'}\rVert_2^2 &\leq \frac{r^2\tilde{\ccalL}^2}{2k^2},
\end{align*}
\end{lemma}

\textit{Proof.}
\begin{align*}
    \lVert \hat{\omega}^{est}_{r} -  \hat{\omega}^{est}_k \otimes  \mathbf{1}_{h' \times h'}\rVert_2^2 &= \sum_{j=1}^{k}\sum_{i=1}^{k} \varepsilon(i,j), \\
\end{align*}
where $\varepsilon(i,j)$ is the error of each $h' \times h'$ boxes and is defined as follows:
\begin{align*}
    \varepsilon(i,j) &= \sum_{jj=\frac{-h'}{2}}^{\frac{h'}{2}}\sum_{ii=\frac{-h'}{2}}^{\frac{h'}{2}} \left( \hat{\omega}^{est}_{r}((i-\frac{1}{2})h' + ii, (j-\frac{1}{2})h' + jj) - \hat{\omega}^{est}_k(i,j) \right)^2, \qquad i,j=1:k \\
\end{align*}
Now, if we consider $\varepsilon_{h'}$ as the upper bound for the above error, we have:
\begin{align*}
    \| \varepsilon(i,j) \| \leq \varepsilon_{h'} \Rightarrow 
    \lVert \hat{\omega}^{est}_{r} -  \hat{\omega}^{est}_k \otimes  \mathbf{1}_{h' \times h'}\rVert_2^2 &\leq k^2 \varepsilon_{h'} \\
\end{align*}
To calculate $\varepsilon_{h'}$, we note that in $\varepsilon_{h'}$, we are summing up the differences of $h'^2$ pixels with the center of a box of size $h' \times h'$. Also, since $f(x)$ is Lipschitz continuous, we can upper bound each difference with the difference of the farthest points from the center, which is the corner of the square. As a result: 

\begin{align*}
    & \| \hat{\omega}^{est}_{r}((i-\frac{1}{2})h' + ii, (j-\frac{1}{2})h' + jj) - \hat{\omega}^{est}_{k}(i,j) \| \leq \\
    & \| \hat{\omega}^{est}_{r}((i-\frac{1}{2})h' + \frac{h'}{2}, (j-\frac{1}{2})h' + \frac{h'}{2}) - \hat{\omega}^{est}_{r}((i-\frac{1}{2})h', (j-\frac{1}{2})h') \|\\
\end{align*}

The above expression holds because

\begin{align*}
    \hat{\omega}^{est}_{r}((i-\frac{1}{2})h', (j-\frac{1}{2})h') &= \hat{\omega}^{est}_{k}(i,j)
\end{align*}

Now, since $\hat{\omega}^{est}_{r}$ is sampled from the continuous function $f(x,y)$, we can calculate the differences in the continuous domain as follows:

\begin{align*}
    \| \hat{\omega}^{est}_{r}((i-\frac{1}{2})h' + \frac{h'}{2}, (j-\frac{1}{2})h' + \frac{h'}{2}) - \hat{\omega}^{est}_{r}((i-\frac{1}{2})h', (j-\frac{1}{2})h') \| = \\
    \|f_\theta(x_i + \frac{1}{2k}, y_j + \frac{1}{2k}) - f_\theta(x_i , y_j) \|_2
\end{align*}

which is true because the domain of the continuous function is $[0,1]$, and we have divided it into $k$ squares, making the size of square $\frac{1}{k} \times \frac{1}{k}$ in the continuous domain.
Based on $f$ being $\tilde{\ccalL}$-Lipschitz, we have:

\begin{align*}
    \|f_\theta(x_i + \frac{1}{2k}, y_j + \frac{1}{2k}) - f_\theta(x_i , y_j) \|_2 &\leq \frac{\tilde{\ccalL}\sqrt{2}}{2k}
\end{align*}

And as a result:
\begin{align*}
    \varepsilon_{h'} &= h'^{2} (\frac{\tilde{\ccalL}\sqrt{2}}{2k}) ^2 \\
\end{align*}
Finally:
 \begin{align*}
    \lVert \hat{\omega}^{est}_{r} -  \hat{\omega}^{est}_k \otimes  \mathbf{1}_{h' \times h'}\rVert_2^2 &\leq k^2 h'^{2} (\frac{\tilde{\ccalL}\sqrt{2}}{2k}) ^2 = \frac{r^2\tilde{\ccalL}^2}{2k^2}.
\end{align*}

\newpage 
\begin{lemma}[MSE of sampled estimation]\label{lemma6}
Let $\omega_r$ denote the sampled true graphon, and assume that the true graphon is Lipschitz continuous with a constant $L>0$. Consider $\hat{\omega}^{est}_{r}$ as the discretization of the function $f_\theta(x,y)$, which is the estimation of the true graphon leaned by SIGL and represented by INR. We can then evaluate the MSE of the sampled estimation as follows:
\begin{align*}
    MSE(\hat{\omega}^{est}_{r}, \omega_r) &\leq \varepsilon_{tr} + 4\ccalL^2\tau^2 + \left(\frac{\tilde{\ccalL}^2}{2} + \frac{C'}{r^2} \right)\frac{1}{k^2} + \left(3+4\ccalL^2\tau^2\right)\frac{k^2}{n^2}
\end{align*}
\end{lemma}
\textit{Proof.}
\begin{align*}
    &MSE(\hat{\omega}^{est}_{r}, \omega_r) \\ 
    &= \frac{1}{r^2} \mathbb{E}\left[\lVert \hat{\omega}^{est}_{r} - \omega_r \rVert_2^2\right] \\
    &= \frac{1}{r^2} \mathbb{E}\left[\lVert \hat{\omega}^{est}_{r} - H^{\omega_r} \otimes \mathbf{1}_{h' \times h'} + H^{\omega_r} \otimes \mathbf{1}_{h' \times h'} -\omega_r \rVert_2^2\right] \\
    &\leq \frac{1}{r^2} \left( \mathbb{E}\left[\lVert \hat{\omega}^{est}_{r} - H^{\omega_r} \otimes \mathbf{1}_{h' \times h'} \rVert_2^2\right] + \mathbb{E}\left[\lVert H^{\omega_r} \otimes \mathbf{1}_{h' \times h'} -\omega_r \rVert_2^2\right] \right) \stackrel{Lemma~\ref{lemma1}}{\Rightarrow}\\
    &\leq \frac{1}{r^2} \left( \mathbb{E}\left[\lVert \hat{\omega}^{est}_{r} - H^{\omega_r} \otimes \mathbf{1}_{h' \times h'} \rVert_2^2\right] + \frac{C'}{k^2} \right) \\
    &\leq \frac{1}{r^2} \left( \mathbb{E}\left[\lVert \hat{\omega}^{est}_{r} - \hat{\omega}^{est}_k \otimes \mathbf{1}_{h' \times h'} + \hat{\omega}^{est}_k \otimes \mathbf{1}_{h' \times h'} - H^{\omega_r} \otimes \mathbf{1}_{h' \times h'} \rVert_2^2\right] + \frac{C'}{k^2} \right) \\
    &\leq \frac{1}{r^2} \left( \mathbb{E}\left[\lVert \hat{\omega}^{est}_{r} - \hat{\omega}^{est}_k \otimes \mathbf{1}_{h' \times h'} \rVert_2^2\right]  + \mathbb{E}\left[\lVert \hat{\omega}^{est}_k \otimes \mathbf{1}_{h' \times h'} - H^{\omega_r} \otimes \mathbf{1}_{h' \times h'} \rVert_2^2\right] + \frac{C'}{k^2} \right) \\
    &\leq \frac{1}{r^2} \left( \mathbb{E}\left[\lVert \hat{\omega}^{est}_{r} - \hat{\omega}^{est}_k \otimes \mathbf{1}_{h' \times h'} \rVert_2^2\right]  + \mathbb{E}\left[ h'^2 \lVert \hat{\omega}^{est}_k - H^{\omega_r} \rVert_2^2\right] + \frac{C'}{k^2} \right)\\
    &\leq \frac{1}{r^2} \left( \mathbb{E}\left[\lVert \hat{\omega}^{est}_{r} - \hat{\omega}^{est}_k \otimes \mathbf{1}_{h' \times h'} \rVert_2^2\right]  + h'^2 \left(\mathbb{E}\left[\lVert \hat{\omega}^{est}_k - \hat{H} \rVert_2^2\right] + \mathbb{E}\left[\lVert \hat{H} - H \rVert_2^2\right] + \mathbb{E}\left[\lVert H - H^{\omega_r} \rVert_2^2\right] \right) + \frac{C'}{k^2} \right) \stackrel{Lemma~\ref{lemma5}}{\Rightarrow}\\
    &\leq \frac{1}{r^2} \left(\frac{r^2\tilde{\ccalL}^2}{2k^2} +  h'^2 \left(\mathbb{E}\left[\lVert \hat{\omega}^{est}_k - \hat{H} \rVert_2^2\right] + \mathbb{E}\left[\lVert \hat{H} - H \rVert_2^2\right] + \mathbb{E}\left[\lVert H - H^{\omega_r} \rVert_2^2\right] \right) + \frac{C'}{k^2} \right) \stackrel{Lemma~\ref{lemma4}}{\Rightarrow}\\
    &\leq \frac{1}{r^2} \left(\frac{r^2\tilde{\ccalL}^2}{2k^2} + h'^2 k^2\varepsilon_{tr} + h'^2 \left(\mathbb{E}\left[\lVert \hat{H} - H \rVert_2^2\right] + \mathbb{E}\left[\lVert H - H^{\omega_r} \rVert_2^2\right] \right) + \frac{C'}{k^2} \right) \stackrel{Lemma~\ref{lemma3}}{\Rightarrow}\\
    &\leq \frac{1}{r^2} \left(\frac{r^2\tilde{\ccalL}^2}{2k^2} + h'^2 k^2\varepsilon_{tr} + h'^2 (\frac{k^4}{n^2} \left(2 + 4\ccalL^2\tau^2\right) + \left(4k^2\ccalL^2\tau^2\right)) + h'^2\mathbb{E}\left[\lVert H - H^{\omega_r} \rVert_2^2\right]+ \frac{C'}{k^2} \right) \stackrel{Lemma~\ref{lemma2}}{\Rightarrow}\\
    &\leq \frac{1}{r^2} \left(\frac{r^2\tilde{\ccalL}^2}{2k^2} + h'^2 k^2\varepsilon_{tr} + h'^2 (\frac{k^4}{n^2} \left(2 + 4\ccalL^2\tau^2\right) + \left(4k^2\ccalL^2\tau^2\right) ) + h'^2\frac{k^4}{n^2}+ \frac{C'}{k^2} \right)  \\
    &\leq \frac{\tilde{\ccalL}^2}{2k^2} + \frac{h'^2 k^2\varepsilon_{tr}}{r^2} + \frac{h'^2}{r^2} (\frac{k^4}{n^2} \left(2 + 4\ccalL^2\tau^2\right) + \left(4k^2\ccalL^2\tau^2\right) ) + \frac{h'^2 k^4}{n^2 r^2} + \frac{C'}{k^2r^2} \stackrel{h'k = r}{\Rightarrow}\\
    &\leq \frac{\tilde{\ccalL}^2}{2k^2} + \frac{k^2\varepsilon_{tr}}{k^2} + \frac{r^2}{r^2k^2} (\frac{k^4}{n^2} \left(2 + 4\ccalL^2\tau^2\right) + \left(4k^2\ccalL^2\tau^2\right) ) + \frac{r^2 k^4}{k^2n^2r^2} + \frac{C'}{k^2r^2} \\
    &\leq \frac{\tilde{\ccalL}^2}{2k^2} + \varepsilon_{tr} + \frac{1}{k^2} (\frac{k^4}{n^2} \left(2 + 4\ccalL^2\tau^2\right) + \left(4k^2\ccalL^2\tau^2\right) ) + \frac{k^2}{n^2} + \frac{C'}{k^2r^2} \\
    &\leq \varepsilon_{tr} + 4\ccalL^2\tau^2 + \left(\frac{\tilde{\ccalL}^2}{2} + \frac{C'}{r^2} \right)\frac{1}{k^2} + \left(3+4\ccalL^2\tau^2\right)\frac{k^2}{n^2}
\end{align*}
\newpage
\subsection{Proposition~\ref{prop.1} proof}

\begin{align*}
    \mathbb{E} [\|f(x,y) - \omega(x,y) \|^2 ]
    &= \mathbb{E} [ \int_0^1 \int_0^1 |f(x,y) - \omega(x,y)|^2 dxdy ] \\ 
    &= \mathbb{E} [ \lim_{\Delta x, \Delta y \rightarrow 0} \sum_i \sum_j |f(i\Delta x,j \Delta y) - \omega(i\Delta x,j \Delta y)|^2 \Delta x \Delta y ] \\
    &= \mathbb{E} [ \lim_{r \rightarrow \infty} \sum_i \sum_j |f(\frac{i}{r},\frac{j}{r}) - \omega(\frac{i}{r},\frac{j}{r})|^2 \frac{1}{r} \frac{1}{r} ] \\
    &= \lim_{r \rightarrow \infty} \mathbb{E} [\frac{1}{r^2} \sum_i \sum_j |f(\frac{i}{r},\frac{j}{r}) - \omega(\frac{i}{r},\frac{j}{r})|^2] \\
    &= \lim_{r \rightarrow \infty} \mathbb{E} [\frac{1}{r^2} \sum_i \sum_j |\hat{\omega}^{est}_{r}(x_i,y_j) - \omega_r(x_i,y_j)|^2] \\
    &= \lim_{r \rightarrow \infty} MSE(\hat{\omega}^{est}_{r}, \omega_r) \stackrel{Lemma~\ref{lemma6}}{\Rightarrow}\\
    &\leq \lim_{r \rightarrow \infty} \varepsilon_{tr} + 4\ccalL^2\tau^2 + \left(\frac{\tilde{\ccalL}^2}{2} + \frac{C'}{r^2} \right)\frac{1}{k^2} + \left(3+4\ccalL^2\tau^2\right)\frac{k^2}{n^2} \\
    &\leq \varepsilon_{tr} + 4\ccalL^2\tau^2 + \left(\frac{\tilde{\ccalL}^2}{2}\right)\frac{1}{k^2} + \left(3+4\ccalL^2\tau^2\right)\frac{k^2}{n^2} \quad \stackrel{k=\frac{n}{h}}{\Rightarrow}\\
    &\leq \varepsilon_{tr} + 4\ccalL^2\tau^2 + \left(\frac{\tilde{\ccalL}^2}{2}\right)\frac{h^2}{n^2} + \left(3+4\ccalL^2\tau^2\right)\frac{1}{h^2} 
\end{align*}

Now since $h<n$ and is an increasing function of $n$ (e.g. $\log n$),
\begin{align*}
     &\lim_{n \rightarrow \infty} \mathbb{E} [\|f(x,y) - \omega(x,y) \|^2 ] \leq \varepsilon_{tr} + 4\ccalL^2\tau^2.
\end{align*}

\vfill

\end{appendix}

\end{document}